**Resolving Prognostic Heterogeneity Among Patients with Identical Illness Severity Scores in Sepsis: A Three-Site Retrospective Study**

Kevin Zhu[1,3], Ryan Zhang[2,3], Baraa Abed[2,3], Tilendra Choudhary, PhD[3,4], Malvern Madondo, PhD[3,4], Mehak Arora[2,3], Yixuan Yang[2,3], Alasdair Gent, PhD[3,4], Aditya Nagori, PhD[3,4], Omer T. Inan, PhD[5,6], Krista L. Haines, DO[3,7], Suresh M. Agarwal, MD[3], Vijay Krishnamoorthy, MD, PhD[4,7], Mihai V. Podgoreanu, MD[4], Joo Heung Yoon, MD[9], Tetsu Ohnuma, MD, PhD, MPH[4], Patrick Georgoff, MD[3], Michael R. Pinsky, MD, CM, Dr hc, FCCP, MCCM[8,9], Gilles Clermont, MD, MSc[8,9], Craig M. Coopersmith, MD, FACS, MCCM[10,11], Craig S. Jabaley, MD, FCCM[10,12], Rishikesan Kamaleswaran, PhD[1,2,3,4]

1 Department of Biomedical Engineering, Duke University, Durham, NC.

2 Department of Electrical and Computer Engineering, Duke University, Durham, NC.

3 Department of Surgery, Duke University School of Medicine, Durham, NC.

4 Department of Anesthesiology, Duke University, Durham, NC.

5 School of Electrical and Computer Engineering, Georgia Institute of Technology, Atlanta, GA.

6 Department of Biomedical Engineering, Georgia Institute of Technology, Atlanta, GA.

7 Department of Population Health Sciences, Duke University School of Medicine, Durham, NC, USA.

8 Department of Critical Care Medicine, University of Pittsburgh, Pittsburgh, PA.

9 School of Medicine, University of Pittsburgh, Pittsburgh, PA.

10 Emory Critical Care Center, Emory Healthcare, Atlanta, GA.

11 Department of Surgery, Emory University School of Medicine, Atlanta, GA.

12 Department of Anesthesiology, Emory University School of Medicine, Atlanta, GA.

*Abstract*

**Objective:** To determine whether an hourly index derived from routinely charted electronic health record data can distinguish mortality risk among sepsis patients with the same Sequential Organ Failure Assessment-2 (SOFA-2) score, and to evaluate this for trial enrichment and crisis triage.

**Design:** A retrospective study with one development cohort and three validation cohorts.

**Setting:** Three academic health systems in Massachusetts, Georgia, and Pennsylvania.

**Patients:** A total of 34,114, 8,733, and 3,779 adult patients meeting Sepsis-3 criteria.

**Interventions:** None.

**Measurements and Main Results:** Using our development cohort, we created an hourly index by averaging 20 neural networks that predicted in-hospital death or hospice discharge from 43 routinely charted EHR variables. We applied it unchanged to all validation cohorts and measured performance with tied-pair AUROC, the probability that a patient who died or went to hospice rated higher risk than a survivor with the same SOFA-2 score (0.5 is chance). At 24 hours after ICU admission, our index scored 0.71-0.79, versus at most 0.65 (95% CI 0.63-0.66) for SOFA-2 plus age, APACHE II, lactate, or change in SOFA-2. In a simulated trial requiring SOFA-2 of at least 6, choosing half of eligible patients by our index needed 27% (95% CI 21%-32%) fewer patients per arm than screening by SOFA-2 alone. In a simulated crisis triage protocol with critical care available for about 80% of patients, using our index instead of SOFA-2 denied critical care to 15.5 (95% CI 13.1-17.9) fewer eventual survivors per 1,000.

**Conclusions:** Routinely charted data identified higher-risk patients among those with the same SOFA-2 score better than existing tools. Screening trial patients by this index instead of SOFA-2 alone could reduce patients needed per arm by about a quarter. The index could also help triage protocols direct critical care toward likely survivors, pending prospective validation.

***Key Points***

**Question:** Can routinely charted data distinguish risk of death among sepsis patients with the same SOFA-2 score?

**Findings:** At three US health systems, an hourly index outperformed existing tools at identifying which of two patients with the same SOFA-2 score died or went to hospice (tied-pair AUROC 0.71-0.79 vs. at most 0.65). In simulations, it reduced trial size by 27% and denied critical care to 15.5 fewer survivors per 1,000 patients in crisis triage.

**Meaning:** Patients with the same SOFA-2 score can differ widely in risk, so decisions based on SOFA-2 may benefit from complementary tools that improve risk assessment, pending prospective validation.

## *Introduction*

Sepsis is life-threatening organ dysfunction caused by a dysregulated host response to infection [1]. It was related to nearly a third of deaths worldwide in 2021 and is often present in hospitalizations that end in death or discharge to hospice [2,3]. Clinicians describe illness severity in sepsis with scores such as the Sequential Organ Failure Assessment (SOFA) [4] and Acute Physiology and Chronic Health Evaluation II (APACHE II) [5]. SOFA and its recent update, SOFA-2, were primarily designed to describe organ dysfunction rather than compete with mortality prediction scores [4,6,7]. However, they are often linked with risk of death, as in the validation of SOFA-2, the Sepsis-3 criteria, and crisis triage protocols [1,6,8].

However, patients with the same score do not necessarily have the same risk of death. Among these patients, mortality has been shown to vary with lactate and with one SOFA version when the other is held constant [9-11]. This matters for uses that treat these patients as having equivalent risk, such as trial enrollment and crisis triage. Clinical trials compare outcomes such as mortality between treatment and control groups. Patients at lower risk of death contribute less information to this comparison because most would survive regardless of treatment. Because a SOFA-based enrollment threshold cannot distinguish risk among patients with the same score, it may include many low-risk patients, making trials longer and more expensive. Biomarkers and machine learning have been proposed for selecting sepsis trial patients [12,13].

Similar concerns arise in crisis triage. When critical care resources are scarce, patients who would survive with treatment may be denied it because a SOFA-based system cannot distinguish them from others in the same priority tier. Protocols may separate these patients by age group, then by lottery [8,14], but a patient more likely to survive may still lose critical care access to one less likely to survive. Such ties matter because many patients share a priority group or fall within a narrow range of scores [6,15].

Previous work has examined parts of this problem. A biomarker and learned phenotypes separated mortality within broader SOFA bands [16,17], validations compared SOFA-2 with the original SOFA [11,18], and simulations compared SOFA-based triage with other protocols and learned allocation policies [14,19-21]. However, none of the studies above reported how well any measure discriminated among patients with the exact same score.

We test whether an index built from routinely charted data can separate patients with sepsis who share a SOFA-2 score by their risk of death or hospice discharge. We develop the index at one health system and evaluate it at three. We measure separation with tied-pair area under the receiver operating characteristic curve (AUROC), which compares only patients with the same SOFA-2 score. We then simulate two uses: trial enrollment and crisis triage.

## ***Materials and Methods***

### *Study Design and Patients*

We conducted a retrospective study at three independent health systems: Medical Information Mart for Intensive Care IV (MIMIC-IV; version 3.1; Beth Israel Deaconess Medical Center, 2008-2022) [22], Emory Healthcare (2015-2022), and University of Pittsburgh Medical Center (2016-2023). We developed our index at MIMIC-IV only, then evaluated it unchanged on held-out MIMIC-IV patients and at the two other sites. We followed the TRIPOD+AI reporting guideline (eTable 1). This study was reviewed and determined to be exempt under Pro00114885 by the Duke University Health System Institutional Review Board.

We included ICU admissions of adults (≥18 years) who met Sepsis-3 criteria. For each admission, we used the ICU stay containing the first ICU hour from 24 hours before to 48 hours after Sepsis-3 onset, setting time zero as the start of that stay. We excluded patients with implausibly high recorded fluids or vasopressors, and patients who died without any recorded fluids or vasopressors, who likely received comfort care only (Appendix 1; eTable 2). We assessed patients still alive in hospital at 1, 6, 12, and 24 hours after ICU admission. 24 hours was used as the main result because it is the usual scoring window for our comparators.

At MIMIC-IV, 20% of admissions with at least 24 hours of ICU data were held out for testing, and the rest were used to develop our index. MIMIC-IV validation patients were those held out plus those with less than 24 hours of data, without anyone who had an admission in the development cohort. All Emory and University of Pittsburgh patients were used for validation. Validation cohorts kept one admission per person (Appendix 1).

*Outcome Definitions*

The primary outcome was in-hospital death or discharge to hospice, which we treated as death because hospice discharge marks a transition to end-of-life care [3]. In sensitivity analyses, we used in-hospital death alone. At MIMIC-IV, we also used all-cause 28- and 90-day mortality.

*Scores Compared*

Our index is the average of 20 neural networks that were trained on MIMIC-IV development patients to predict in-hospital death or hospice discharge from a single hour of data. Each hour was described by 43 routinely charted vital signs, laboratory values, and demographic variables, with no inputs for vasopressors, fluids, or dialysis (eTable 3). Values were truncated to plausible ranges [23], and missing values were first filled with the patient's last value, then the development cohort median (Appendix 2).

We compared our index with SOFA-2, as well as APACHE II; SOFA-2 plus age, which separates patients with the same SOFA-2 by age; the highest lactate since 6 hours before ICU admission; and the change in SOFA-2 since the first ICU hour. These were chosen since they are methods that a clinician might use to separate patients with the same SOFA-2 score. SOFA-2 and APACHE II used the worst values from 6 hours before ICU admission to each time, so that tests done before transfer to the ICU were accounted for. SOFA-2 omitted rules that needed unavailable data (Appendix 3). As negative controls, we trained 40 networks of the same design on randomly assigned or shuffled outcome labels, which show how such a model performs by chance.

*Evaluation*

**Separating patients with the same SOFA-2 score.** Within each SOFA-2 value, we compared risk of death or hospice discharge across age groups and across thirds of our index, a simple but arbitrary split that was set on development patients. We considered a difference clear when 95% CIs did not overlap. These comparisons show whether risk differs among patients with the same SOFA-2 score, but not how well a score separates them by risk. Our main measure was therefore tied-pair AUROC, the probability that a patient who died or went to hospice was rated as higher risk than a survivor with the same SOFA-2 score. By definition, SOFA-2 has a tied-pair AUROC of 0.5. We repeated this at each time, with versions of our index with age fixed at 65 or blind to age, sex, and weight, and within female, male, Black, and White patients.

**Simulated trial enrollment.** Trials that enroll higher-risk patients can be smaller, because the same relative effect prevents more deaths. We simulated trials that enrolled patients in two steps. First, patients were eligible if their SOFA-2 was at least 6, a boundary between triage bands that we chose as an example. Second, the trial enrolled a share of the eligible patients, starting with the sickest. This was determined by SOFA-2 alone, by SOFA-2 with ties broken by our index, by our index alone, or by APACHE II, with ties at the cutoff broken at random. Because there is no standard share of eligible patients to enroll, we used 10% to 90%, using 50% for the main results. Following an earlier study [12], we assumed a 20% relative reduction in death or discharge to hospice compared to the control arm. We then calculated the number of patients required per arm to detect this difference with 80% power and two-sided $\alpha$ of 0.05 (Appendix 4). We report the ratio to screening by SOFA-2 alone. With half of eligible patients enrolled, we repeated this at earlier times, with other outcomes, with a 15% reduction, and after removing patients who died or went to hospice within 24 hours of enrollment, because a treatment would have had little time to help.

**Simulated crisis triage.** We simulated a model hospital policy for allocating scarce critical care resources (March 26, 2020 version) [8]. It places patients in four SOFA bands (<6, 6-8, 9-11, and ≥12) in order of likelihood of survival. If maximum capacity is reached within a band, patients are grouped by age and treated from youngest to oldest; if capacity is reached within an age group, a lottery is used. We set hospital capacity as the number of patients with SOFA-2 below 12, 9, or 6, corresponding to mild, moderate, and severe shortages, since the SOFA-2 protocol does not need tie-breakers at these levels. We compared using only SOFA-2 to set the severity bands with using SOFA-2 plus age, APACHE II, and a version of our index blind to age, sex, and weight. We used the blind version because the policy already considers age, and because sex, and weight as a proxy for it, should not decide who receives care. For the other scores, we calculated severity band cutoff-points such that each band had the same share of MIMIC-IV development patients as its SOFA-2 analogue, and applied them unchanged at all sites (Appendix 5).

Because every patient received ICU care, we could not measure lives saved. We instead counted survivors excluded: patients left without critical care who survived to hospital discharge, not to hospice, per 1,000 patients. At the moderate shortage, we repeated this at 6 hours, at a 72-hour reassessment, with other cutoff-points, with a lottery as the only tie-breaker, and by sex and race.

*Statistical analysis*

Outcome rates have Wilson 95% CIs. Other intervals came from 2,000 bootstrap resamples of patients, shared by all scores so that differences are paired. We pooled sites with random-effects models, which allow the gain to differ between hospitals, and calculated 95% prediction intervals, the range expected at a new hospital. We also measured how often decisions changed between adjacent hours (Appendix 6). Analyses used Python 3.11 and PyTorch 2.9. Claude Code (Anthropic) and Codex (OpenAI) helped write analysis code and text. The authors verified all output and take responsibility for it.

## ***Results***

### *Patient Characteristics*

The development cohort included 23,293 admissions from MIMIC-IV. Our index was then tested on 10,821 MIMIC-IV, 8,733 Emory, and 3,779 University of Pittsburgh patients (Table 1).

### *Separating Risk Within Patients with Equal SOFA-2 Scores*

Patients with the same SOFA-2 score differed in their risk of death or hospice discharge. Those over 75 years had clearly higher risk than those aged 18 to 40 at 11 of 15 SOFA-2 values (Figure 1A). Our index separated patients further: at every SOFA-2 value and site, risk was higher in its highest third than in its lowest (Figure 1B-D). Some patients with low SOFA-2 scores had risk as high as or higher than that of patients with much higher scores. At MIMIC-IV, risk was 17.2% in the highest third at SOFA-2=3 and 8.5% in the lowest third at SOFA-2=11.

At 24 hours, no existing tool had a tied-pair AUROC above 0.65 at any site, while our index reached 0.71 to 0.79 (Figure 2A-B; eTable 4). Calibration for all scores showed predicted risk was too low at Emory and University of Pittsburgh (eTable 5; eFigure 1). Lactate and change in SOFA-2 fell within the range of the negative controls at every site, as did APACHE II at Emory and University of Pittsburgh. Pooled across sites, our index exceeded SOFA-2 plus age, APACHE II, and lactate by 0.119 to 0.206 (lowest 95% CI bound 0.049).

These differences held at 1, 6, and 12 hours (at least 0.084; lowest 95% CI bound 0.038), with our index blind to age, sex, and weight (at least 0.085; lowest bound 0.008), and within Black, White, female, and male patients (at least 0.104; every lower bound above 0; Figure 2C). Their size varied between sites, especially against SOFA-2 plus age and lactate, but always favored our index (eTable 6).

*Simulated Trial Enrollment*

Screening patients with our index gave a smaller trial with the same power (Figure 3A). Compared with screening by SOFA-2 alone, it required 0.50 (95% CI 0.44-0.57) times the patients per arm with 10% of eligible patients enrolled and 0.94 (0.92-0.95) times with 90%. At 50%, the ratio was 0.73 (0.68-0.79), compared with 0.94 (0.92-0.97) for breaking SOFA-2 ties with our index and 0.99 (0.95-1.03) for APACHE II.

This gain held at 1, 6, and 12 hours, with death alone as the endpoint, with a 15% treatment effect, and for 28- and 90-day mortality at MIMIC-IV (ratios 0.61 to 0.83; Figure 3B-C). It was not driven by patients who die too soon to benefit, because the ratio stayed at 0.73 without patients who died or went to hospice within 24 hours (Figure 3D). Compared with screening by SOFA-2 alone, our index chose older patients and fewer receiving mechanical ventilation, and the two methods shared 60% to 65% of enrolled patients (eTable 7).

*Simulated Crisis Triage*

A protocol using our index excluded fewer survivors than one using SOFA-2 at every shortage level and site (Figure 4A). At the moderate shortage, it excluded 15.5 (95% CI 13.1-17.9) fewer survivors per 1,000 patients than SOFA-2, 11.4 (7.6-15.2) fewer than SOFA-2 plus age, and 17.4 (14.6-20.2) fewer than APACHE II, pooled across sites.

The gain over SOFA-2 held at 6 hours, at a 72-hour reassessment, and with other cutoff-points (10.6 to 15.8 fewer survivors excluded; Figure 4B-C). It did not widen pooled gaps between groups (Figure 4D). Per 1,000 patients in each group, the SOFA-2 protocol excluded 28.5 fewer female than male survivors, compared with 0.9 more under our index, and 13.0 more Black than White survivors, compared with 4.0 more. At MIMIC-IV alone, the Black-White gap was larger under our index (16.5, compared with 8.9; difference 7.6, 95% CI -15.3 to 31.5). At University of Pittsburgh, our index excluded 168 (95% CI 120-193) more survivors per 1,000 patients over age 75 than SOFA-2, but the age tie-breaker caused this. With a lottery alone, the difference was 12 (-13 to 37) (eTable 8).

*Stability*

Between adjacent hours, decisions changed for 2.7% to 3.9% of patients with our index and 0.5% to 0.6% with SOFA-2. Averaging our index over 3 hours about halved this (eTable 9).

## *Discussion*

This study has three main findings. First, an hourly index built from routinely charted data separated risk among patients with the same SOFA-2 score better than existing tools at all three health systems. Second, in simulated trials, choosing half of eligible patients with our index needed 27% fewer patients per arm than screening by SOFA-2 alone. Third, in a simulated crisis triage protocol, using our index at a moderate shortage excluded 11 to 17 fewer survivors per 1,000 patients than the other scores.

Earlier studies found mortality differences among patients with the same SOFA score [9-11], but did not compare how well clinical tools distinguish risk within these groups. Such differences show that risk varies among patients with the same SOFA-2 score, but not how well a score separates them by risk. For example, patients with lactate above 4 mmol/L clearly had worse outcomes than those with lactate of 2 or lower at 7 of 15 SOFA-2 values (eTable 10), yet lactate separated patients with the same SOFA-2 score little better than chance (tied-pair AUROC 0.52 to 0.55). The change in SOFA-2 since ICU admission did no better than negative controls. APACHE II and SOFA-2 plus age did better (pooled tied-pair AUROC 0.60 and 0.62), but our index still exceeded SOFA-2 plus age with its own age input fixed at 65, so its gain did not come from age alone.

These results also have practical uses. For example, trials may enroll higher-risk patients with biomarkers to reduce sample size [12]. Unlike biomarkers, our index uses data that are routinely charted. Using it only to break SOFA-2 ties gave a smaller gain, most likely because different SOFA-2 values can have overlapping mortality risks (Figure 1). However, a trial using our index would enroll a different population, so trialists would need to consider whether treatment works as well in these patients.

Another application could be crisis triage, where the policy simulated here allows SOFA to be replaced by another validated measure of survival [8]. Our index had a smaller pooled gap by sex than SOFA-2, and the pooled Black-White gap was not clearly different, although it was larger at MIMIC-IV. This matters because SOFA has overestimated mortality in Black patients, which could deprioritize them during crises [24]. Other simulations of ventilator allocation reached mixed conclusions [19,20], and the SOFA-2 developers ask that SOFA use for triage be validated first [7]. Only prospective evaluation can settle whether any replacement score improves triage.

This study has several limitations. It used retrospective data from three US health systems. Due to data availability, SOFA-2 omitted the rules for urine output, sedation and delirium, extracorporeal membrane oxygenation, and mechanical circulatory support. APACHE II's chronic health points came from diagnosis codes that were recorded more fully at Emory than elsewhere, and other ICU scores were not compared. Both simulations assumed simplified conditions: the trial assumed that a treatment works equally well at every level of risk, the triage protocol omitted comorbidity points, and neither use was tested prospectively. The size of our index's advantages varied between sites, and prediction intervals for a new hospital sometimes included no gain (eTable 6), so reported results could be smaller or absent. Finally, 28- and 90-day mortality were available only at MIMIC-IV, and there were not enough Hispanic and Asian patients for subgroup analyses.

***Conclusions***

Patients with sepsis who share a SOFA-2 score can differ widely in their risk of death, and an index built from routinely charted data separated them by risk better than existing tools. Used alongside SOFA-2, it could make sepsis trials smaller and help triage protocols identify likely survivors. Future work includes prospective testing, and ensuring stability in its hourly decisions.

***References***

[1] Singer, M. *et al.* The third international consensus definitions for sepsis and septic shock (sepsis-3). *JAMA* **315**, 801-810 (2016).

[2] Gray, A. P. *et al.* Global, regional, and national sepsis incidence and mortality, 1990-2021: a systematic analysis. *The Lancet Global Health* **13**, e2013-e2026 (2025).

[3] Rhee, C. *et al.* Prevalence, underlying causes, and preventability of sepsis-associated mortality in US acute care hospitals. *JAMA Network Open* **2**, e187571 (2019).

[4] Vincent, J.-L. *et al.* The SOFA (sepsis-related organ failure assessment) score to describe organ dysfunction/failure: On behalf of the working group on sepsis-related problems of the European Society of Intensive Care Medicine. *Intensive Care Medicine* **22**, 707-710 (1996).

[5] Knaus, W. A., Draper, E. A., Wagner, D. P. & Zimmerman, J. E. APACHE II: a severity of disease classification system. *Critical Care Medicine* **13**, 818-829 (1985).

[6] Ranzani, O. T. *et al.* Development and validation of the sequential organ failure assessment (SOFA)-2 score. *JAMA* **334**, 2090-2103 (2025).

[7] Moreno, R. *et al.* Rationale and methodological approach underlying the development of the sequential organ failure assessment (SOFA)-2 score: a consensus statement. *JAMA Network Open* **8**, e2545040 (2025).

[8] White, D. B. & Lo, B. A framework for rationing ventilators and critical care beds during the COVID-19 pandemic. *JAMA* **323**, 1773-1774 (2020).

[9] Ono, S., Uchino, S., Kumaido, Y., Yawata, S. & Katayama, S. Patient-level reclassification and discordance between SOFA and SOFA-2: a multicenter ICU study. *Critical Care* **30**, 425 (2026).

[10] Park, H. *et al.* Serial evaluation of the serum lactate level with the sofa score to predict mortality in patients with sepsis. *Scientific Reports* **13**, 6351 (2023).

[11] Bai, S. *et al.* SOFA-2 versus SOFA-1 for mortality prediction in critically ill patients. *Journal of Intensive Care* **14**, 32 (2026).

[12] Anderson, B. J. *et al.* Plasma sTNFR1 and IL8 for prognostic enrichment in sepsis trials: a prospective cohort study. *Critical Care* **23**, 400 (2019).

[13] Yang, M. *et al.* Enhancing patient selection in sepsis clinical trials design through an AI enrichment strategy: Algorithm development and validation. *Journal of Medical Internet Research* **26**, e54621 (2024).

[14] Bhavani, S. V. *et al.* Simulation of ventilator allocation in critically ill patients with covid-19. *American Journal of Respiratory and Critical Care Medicine* **204**, 1224-1227 (2021).

[15] Wunsch, H. *et al.* Comparison of 2 triage scoring guidelines for allocation of mechanical ventilators. *JAMA Network Open* **3**, e2029250 (2020).

[16] Andaluz-Ojeda, D. *et al.* Superior accuracy of mid-regional proadrenomedullin for mortality prediction in sepsis with varying levels of illness severity. *Annals of Intensive are* **7**, 15 (2017).

[17] Feng, X. *et al.* Subphenotyping sepsis based on organ interaction trajectory using a deep temporal graph clustering model: a retrospective cohort study. *EClinicalMedicine* **90** (2025).

[18] Helleberg, J., Sundelin, A., Soltani, N., Rooyackers, O. & Mårtensson, J. Predictive validity of daily sequential organ failure assessment (sofa)-2 score for 30-day mortality. *Critical Care* **30**, 280 (2026).

[19] Herington, J. *et al.* Investigating ethical tradeoffs in crisis standards of care through simulation of ventilator allocation protocols. *PLoS One* **19**, e0300951 (2024).

[20] Grand-Clément, J., Goh, Y. H., Chan, C., Goyal, V. & Chuang, E. Interpretable machine learning for resource allocation with application to ventilator triage. *arXiv preprint arXiv:2110.10994* (2021).

[21] Li, Y. *et al.* Deep reinforcement learning for efficient and fair allocation of health care resources. *Proceedings of the Thirty-Fourth International Joint Conference on Artificial Intelligence, IJCAI-25*, 9790-9798 (2025).

[22] Johnson, A. E. *et al.* MIMIC-IV, a freely accessible electronic health record dataset. *Scientific Data* **10**, 1 (2023).

[23] Arora, M. *et al.* Improving clinical decision support through interpretable machine learning and error handling in electronic health records. *Journal of the American Medical Informatics Association* **33**, 123-132 (2026).

[24] Miller, W. D., Han, X., Peek, M. E., Charan Ashana, D. & Parker, W. F. Accuracy of the sequential organ failure assessment score for in-hospital mortality by race and relevance to crisis standards of care. *JAMA Network Open* **4**, e2113891 (2021).

*Acknowledgements*

**Funding:** KZ, RZ, BA, TC, MM, MA, YY, AG, AN, OTI, MRP, GC, CMC, and RK were supported by R01GM139967. The funders had no role in the design, analysis, or reporting of the study.

**Conflicts of interest:** The authors declare no conflicts of interest.

**Computing resources:** Analyses were run on the Duke Compute Cluster at Duke University.

**Protocol, registration, and patient involvement:** No protocol was published. The study was not registered. Patients and the public were not involved in the study.

**Data and code availability:** The code for all analyses will be available at https://github.com/Kamaleswaran-Lab/sofa2-risk-separation. MIMIC-IV is available to credentialed users through PhysioNet [22]. Emory Healthcare and University of Pittsburgh Medical Center data cannot be shared because of patient privacy restrictions.

***Figures and Tables***

**Table 1:** Patient Characteristics

| **Characteristic** | MIMIC-IV development | MIMIC-IV validation | Emory | University of Pittsburgh |
|---|---|---|---|---|
| **Patients** | | | | |
| Alive in hospital at 1 h, n | 23,293 | 10,821 | 8,733 | 3,779 |
| Alive in hospital at 6 h, n | 23,293 | 10,779 | 8,709 | 3,773 |
| Alive in hospital at 12 h, n | 23,293 | 10,661 | 8,670 | 3,761 |
| Alive in hospital at 24 h, n | 23,267 | 10,349 | 8,523 | 3,716 |
| **Demographics** | | | | |
| Age (y), median [IQR] | 65 [54, 76] | 67 [56, 78] | 64 [52, 73] | 62 [51, 72] |
| Age category (y), n (%) | | | | |
| 18–40 | 2,206 (9.5) | 828 (8.0) | 1,013 (11.9) | 533 (14.3) |
| 41–60 | 6,782 (29.1) | 2,754 (26.6) | 2,573 (30.2) | 1,127 (30.3) |
| 61–75 | 8,132 (35.0) | 3,728 (36.0) | 3,171 (37.2) | 1,406 (37.8) |
| >75 | 6,147 (26.4) | 3,039 (29.4) | 1,766 (20.7) | 650 (17.5) |
| Female, n (%) | 9,993 (42.9) | 4,306 (41.6) | 3,942 (46.3) | 1,605 (43.2) |
| Race and ethnicity, n (%) | | | | |
| Hispanic of any race | 895 (3.8) | 360 (3.5) | — | 17 (0.5) |
| Black | 2,436 (10.5) | 872 (8.4) | 3,711 (43.5) | 470 (12.6) |
| White | 15,559 (66.9) | 7,141 (69.0) | 4,000 (46.9) | 2,791 (75.1) |
| Asian | 694 (3.0) | 306 (3.0) | 283 (3.3) | 25 (0.7) |
| Other or unknown | 3,683 (15.8) | 1,670 (16.1) | 529 (6.2) | 413 (11.1) |
| Weight (kg), median [IQR] | 79.7 [66.5, 95.0] | 79.9 [67.5, 95.0] | 79.4 [65.5, 97.1] | 83.4 [68.1, 102.0] |
| **Severity at 24 h** | | | | |
| SOFA-2, median [IQR] | 5 [3, 7] | 4 [3, 6] | 5 [3, 8] | 6 [4, 9] |
| SOFA-2 group, n (%) | | | | |
| <6 | 13,618 (58.5) | 6,741 (65.1) | 4,605 (54.0) | 1,608 (43.3) |
| 6–8 | 5,985 (25.7) | 2,440 (23.6) | 2,123 (24.9) | 1,106 (29.8) |
| 9–11 | 2,461 (10.6) | 823 (8.0) | 1,111 (13.0) | 622 (16.7) |
| ≥12 | 1,203 (5.2) | 345 (3.3) | 684 (8.0) | 380 (10.2) |
| APACHE II, median [IQR] | 19 [15, 24] | 18 [14, 23] | 21 [16, 26] | 22 [17, 28] |
| Any APACHE II chronic health points, n (%) | 7,019 (30.2) | 2,371 (22.9) | 4,792 (56.2) | 1,366 (36.8) |
| Lactate measured, n (%) | 17,503 (75.2) | 7,726 (74.7) | 5,671 (66.5) | 3,021 (81.3) |
| Highest lactate (mmol/L), median [IQR] | 2.1 [1.4, 3.4] | 2.2 [1.5, 3.3] | 2.1 [1.3, 3.8] | 2.0 [1.3, 3.5] |
| Lactate group (mmol/L), n (%) | | | | |
| ≤2.0 | 8,385 (36.0) | 3,471 (33.5) | 2,741 (32.2) | 1,551 (41.7) |
| >2.0 to 4.0 | 6,038 (26.0) | 3,013 (29.1) | 1,620 (19.0) | 850 (22.9) |
| >4.0 | 3,080 (13.2) | 1,242 (12.0) | 1,310 (15.4) | 620 (16.7) |
| Not measured | 5,764 (24.8) | 2,623 (25.3) | 2,852 (33.5) | 695 (18.7) |
| **Organ support to 24 h** | | | | |
| Invasive mechanical ventilation, n (%) | 12,149 (52.2) | 4,852 (46.9) | 2,800 (32.9) | 2,135 (57.5) |
| Vasopressors, n (%) | 9,201 (39.5) | 3,834 (37.0) | 2,896 (34.0) | 1,421 (38.2) |
| Kidney replacement therapy, n (%) | 1,394 (6.0) | 341 (3.3) | 537 (6.3) | 216 (5.8) |
| **Outcomes** | | | | |
| In-hospital death or hospice discharge, n (%) | 3,497 (15.0) | 1,160 (11.2) | 1,974 (23.2) | 932 (25.1) |
| In-hospital death, n (%) | 2,966 (12.7) | 996 (9.6) | 1,178 (13.8) | 807 (21.7) |
| Hospice discharge, n (%) | 531 (2.3) | 164 (1.6) | 796 (9.3) | 125 (3.4) |
| Death or hospice discharge within 24 h after the landmark, n (%) | 372 (1.6) | 99 (1.0) | 189 (2.2) | 69 (1.9) |
| ICU length of stay (d), median [IQR] | 3.1 [1.8, 6.2] | 1.9 [1.0, 3.9] | 3.9 [2.0, 8.5] | 4.5 [2.3, 9.0] |
| Hospital length of stay (d), median [IQR] | 9.1 [5.6, 15.9] | 8.0 [5.3, 13.7] | 10.3 [5.9, 18.3] | 14.0 [7.9, 25.0] |
| Death within 28 d, n (%) | 3,753 (16.1) | 1,277 (12.3) | — | — |
| Death within 90 d, n (%) | 5,442 (23.4) | 1,904 (18.4) | — | — |
| **Analysis** | | | | |
| Death–survivor pairs with the same SOFA-2, millions | 5.41 | 0.91 | 0.94 | 0.18 |

IQR, interquartile range; SOFA-2, Sequential Organ Failure Assessment 2; APACHE II, Acute Physiology and Chronic Health Evaluation II. Patients were adults with Sepsis-3 who were alive in hospital 24 hours after ICU admission. We used the ICU stay that contained the first ICU hour between 24 hours before and 48 hours after sepsis onset. MIMIC-IV development patients trained our index and set the severity band cutoff-points. They also had to have at least 24 hours of data in the study window, a rule that the validation patients did not have to meet. Validation patients at MIMIC-IV had no admission in the development cohort, and each validation cohort kept one admission per person. The rows for 1, 6, and 12 hours count patients alive in hospital at those times, including patients who died or left before 24 hours. Severity rows use the worst values from 6 hours before ICU admission to 24 hours. Weight is the last value charted in that window, and organ support is any use in that window. Highest lactate is among patients with a lactate in that window. Due to data availability, SOFA-2 omits the rules for urine output, sedation and delirium, ECMO, and mechanical circulatory support; missing values were scored as normal. APACHE II chronic health points come from diagnosis codes in the current and earlier admissions. Race and ethnicity are as recorded at each site. Ethnicity was unable to be extracted at Emory, so its Black and White groups may include Hispanic patients. ICU length of stay is the stay that contained ICU admission. Death within 28 and 90 days is counted from the 24-hour landmark and is available only at MIMIC-IV.

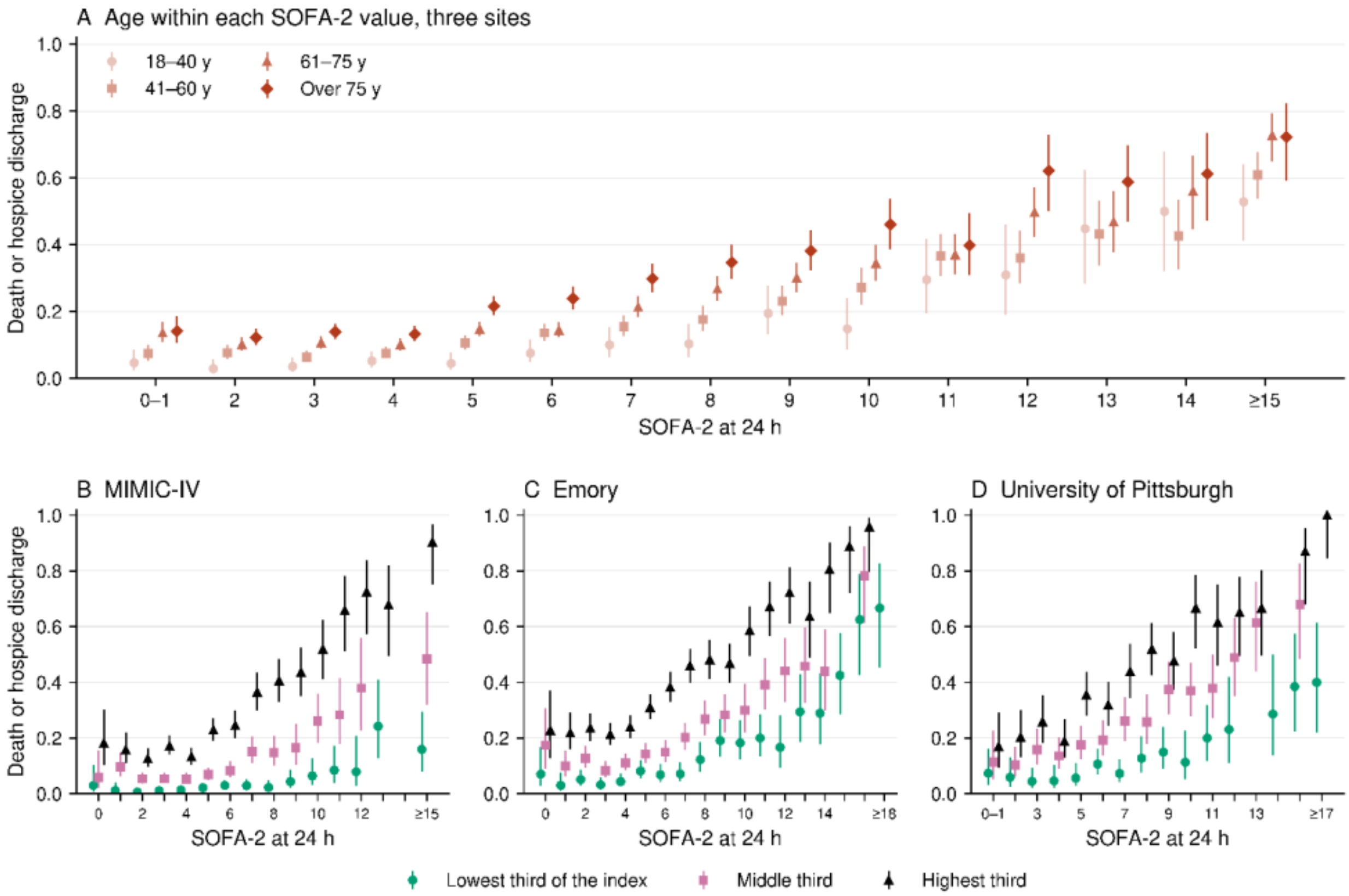

**Figure 1. Patients with the same SOFA-2 have different risks of death.** Death or hospice discharge at each SOFA-2 value at 24 hours after ICU admission, with Wilson 95% intervals. (A) The three sites combined, by age group (18-40, 41-60, 61-75, and over 75). (B-D) MIMIC-IV, Emory, and University of Pittsburgh, by thirds of our index. The thirds were set within each SOFA-2 value on MIMIC-IV development patients, then fixed, and every patient shown was held out from training. High SOFA-2 values were merged until each group had at least 50 patients and 10 deaths at that site; alternate values are labelled. A group with fewer than 20 patients at a value is not plotted.

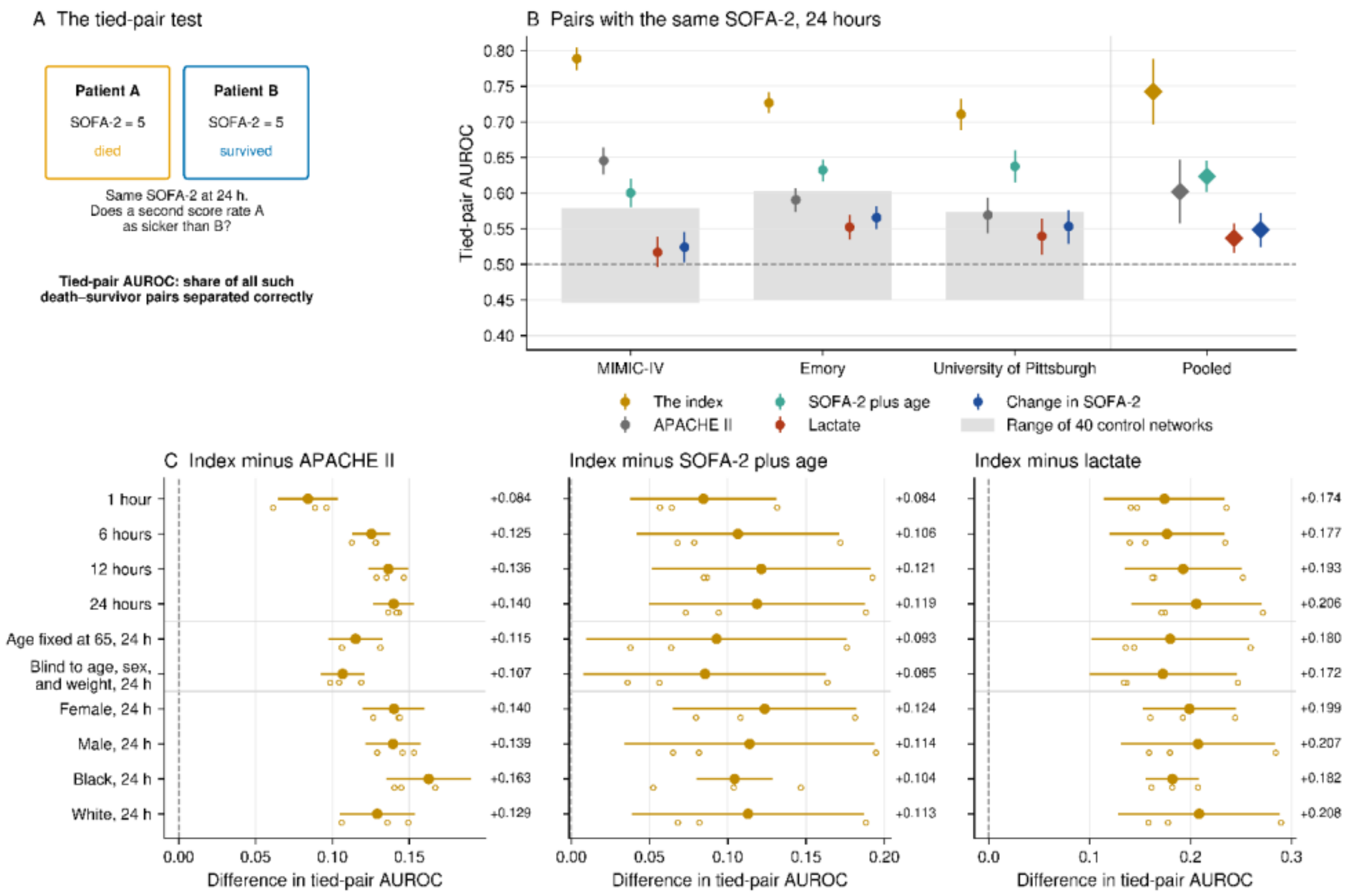

**Figure 2. Existing tools separate these patients poorly, and our index separates them better.** (A) The tied-pair AUROC measures how well a score distinguishes patients with the same SOFA-2 score but different outcomes. It considers all pairs in which one patient died or was discharged to hospice and the other survived, and measures how often the score assigns lower severity to the survivor. Because SOFA-2 assigns the same value to both patients, a score based solely on SOFA-2 has an AUROC of 0.5. (B) Tied-pair AUROC at 24 hours with 95% intervals. Within a SOFA-2 tie, SOFA-2 plus age separates patients by age alone. Lactate is the highest value from 6 hours before ICU admission to 24 hours; a patient with no lactate is rated as lower risk than every patient with one, and two such patients count as half a correct pair. The change in SOFA-2 is the current-hour value minus its value at hour 1. Grey bands show the range of negative control networks at each site, which shows what a model of this form reaches by chance. (C) Our index minus each comparator. Filled circles are pooled with random effects, with 95% intervals, and the small open circles below them are the single sites. Rows show each landmark hour; our index with age fixed at 65; our index blind to age, sex, and weight; and subgroups, in which both patients in each pair belong to the group. Intervals are from 2,000 paired patient resamples. At Emory, ethnicity was not extracted, so its Black and White groups may include Hispanic patients.

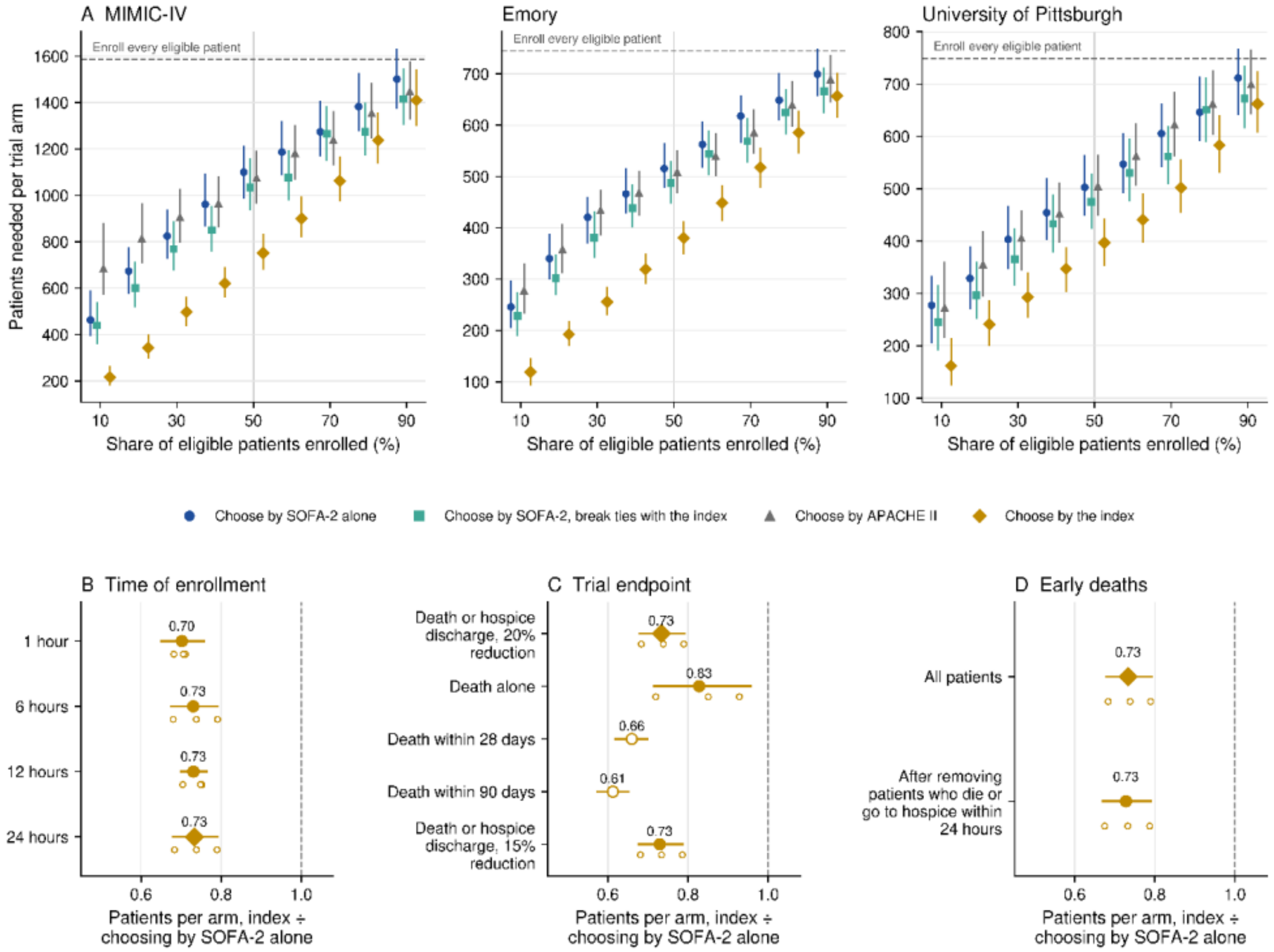

**Figure 3. Screening patients by our index gives a smaller trial with the same power.** We simulate a trial that requires SOFA-2 of at least 6 and enrolls eligible patients by: (1) screening by highest SOFA-2 score alone, (2) breaking SOFA-2 ties with our index, (3) screening with our index only, or (4) screening with APACHE II only. We assume a 20% relative reduction in death or hospice discharge, and require 80% power and two-sided α of 0.05. Fewer patients are needed because our index enrolls patients more likely to die or go to hospice without treatment. (A) Patients per arm by the share of eligible patients enrolled, at 24 hours, with 95% bootstrap intervals, shown for every 10% of eligible patients. Points at the same share are offset for readability. Ties at the cutoff are broken at random. The vertical line marks the 50% share used in panels B-D. (B-D) Patients per arm when screening by our index, divided by patients per arm when screening by highest SOFA-2 score alone, with 50% of eligible patients enrolled. A value below 1 means a smaller trial. Filled circles are pooled across sites with 95% intervals. The diamond is the same pooled estimate for the main analysis. The small open circles below each estimate are the single sites. (B) Each time point only includes patients still alive in hospital at that time. (C) Death alone counts hospice discharge as survival. The 28- and 90-day rows use MIMIC-IV only due to data availability. (D) Patients who died or went to hospice within 24 hours were removed before enrollment. Before removal, 5.1% of enrolled patients died or went to hospice within 24 hours when screening by highest SOFA-2 score alone, and 5.8% when screening by our index.

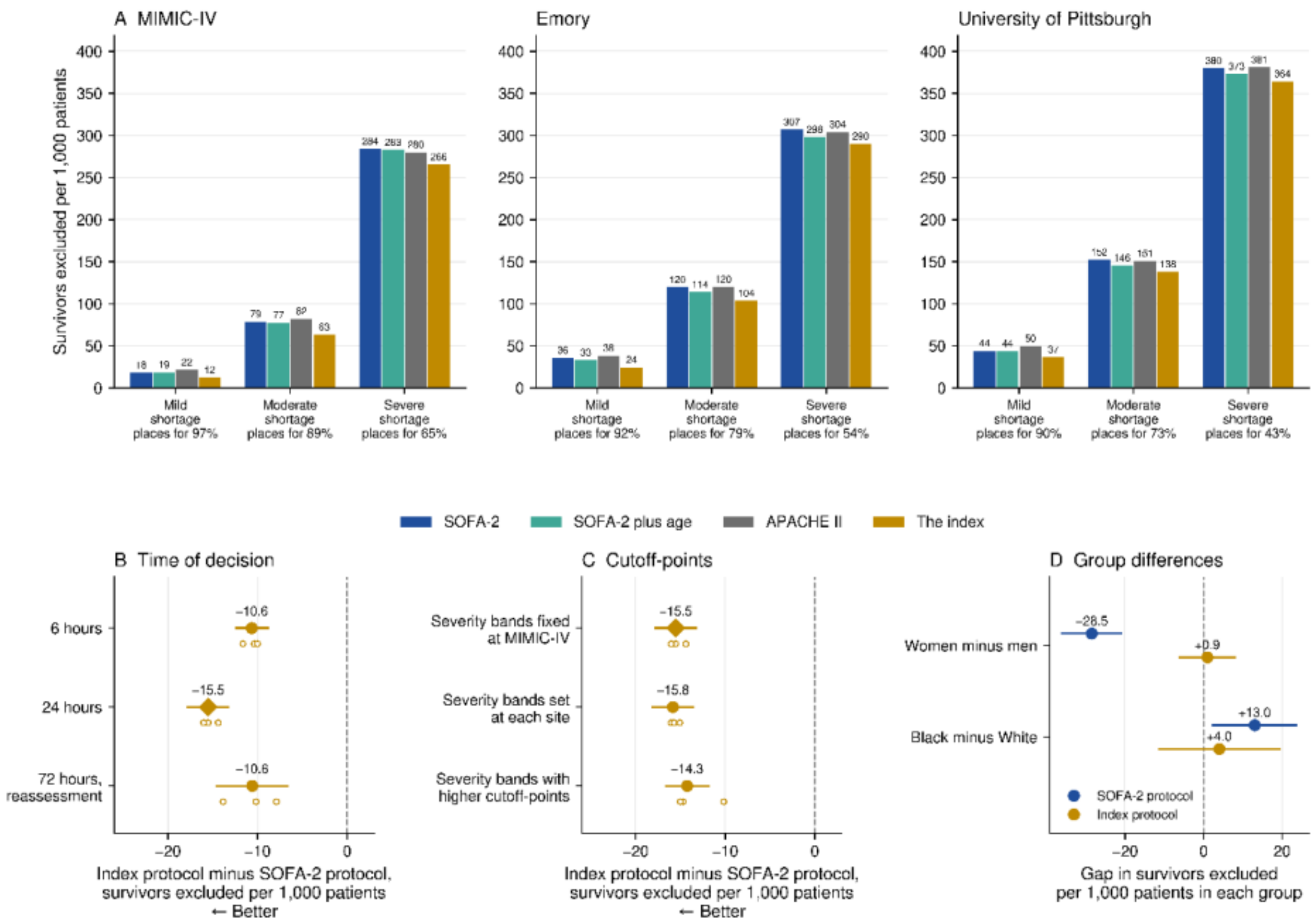

**Figure 4. Replacing SOFA-2 in a triage protocol excludes fewer survivors.** Survivors excluded are patients whom a protocol would leave without critical care, but who survived to discharge (not to hospice) with the ICU care that they received, per 1,000 patients (fewer is better). This shows how well each protocol identifies who is likely to survive, not how many lives it would save. Capacity was set at the number of patients with SOFA-2 below 12 (mild shortage), 9 (moderate), or 6 (severe). Each protocol selected the same number of patients using its respective severity score. To ensure a fair comparison, each replacement score was divided into four severity bands, with band cutoff-points determined using MIMIC-IV development patients so that each band contained the same proportion of patients as the corresponding SOFA-2 band. Within the last severity band served, patients were chosen first by age category, and then at random. No additional points were assigned for comorbidities. The index used here is blind to age, sex, and weight. (A) Survivors excluded by protocol, shortage, and site at 24 hours. (B-C) Our index protocol minus the SOFA-2 protocol at the moderate shortage, where below 0 means fewer survivors excluded. Filled circles are pooled with 95% intervals. The diamond is the same pooled estimate for the main analysis. The small open circles below each estimate are the single sites. (B) Severity bands were recalculated at each landmark hour. (C) The higher cutoff-points divide the index such that its bands hold the same shares of patients as SOFA-2 ≤8, 9-11, 12-14, and >14. (D) The gap between groups in survivors excluded per 1,000 patients in each group, at moderate shortage. The dashed line marks parity; closer to 0 is more equal treatment. At Emory, ethnicity was not extracted, so its Black and White groups may include Hispanic patients.

**Supplemental Content**

## Appendix 1: Cohort Construction

*Sepsis-3 onset.* Sepsis-3 onset was computed as the earlier of a time of suspected infection and a time of qualifying organ dysfunction. We defined suspected infection as either a culture ordered within 24 hours after an anti-infective order, or an anti-infective ordered within 72 hours after a culture. Suspicion time was taken as the earlier time of each pair. Organ dysfunction was defined as the first hour showing a ≥2-point increase in the total SOFA score over a 24-hour rolling window, searched from 24 hours before to 12 hours after each suspicion time. Sepsis-3 time was then the minimum of the suspicion and organ-dysfunction times.

*Inclusion and ICU stay.* We included adults (18 years or older) with a Sepsis-3 onset time and an ICU stay. We used the first continuous ICU stay with any hour from 24 hours before to 48 hours after onset. ICU admission was the true start of that stay, even if it began before the window.

*Exclusions.* We excluded patients with no ICU hour in that window, patients with an average of 6,000 mL or more of fluids per hour or a vasopressor dose of 100 or more norepinephrine equivalents (micrograms per kilogram per minute, with vasopressin in units per minute; weights: norepinephrine 1, epinephrine 1, dopamine 0.01, phenylephrine 0.1, and vasopressin 2.5), and patients who died or went to hospice with no recorded fluids or vasopressors during the admission. Counts are in eTable 2.

*Outcome.* In-hospital death and hospice discharge were taken from the recorded discharge disposition, with the same rule for every patient.

*Development and validation cohorts.* At MIMIC-IV, 20% of admissions with at least 24 hours of data were held out, stratified by outcome. The other admissions were divided into four folds by person, stratified by outcome (seed 42), and used as the development cohort.

To create the validation cohort, we first combined the held-out admissions with all admissions that had less than 24 hours of data. We removed patients with admissions in the development cohort, and took the first admission for those with multiple.

At Emory and University of Pittsburgh, all patients were used for validation, with one admission per person, keeping the first admission for both.

*Sample size.* No sample size calculation was done. All eligible patients were used. The precision of each estimate is shown by its confidence interval.

## Appendix 2: Index Construction

*Inputs.* The inputs were chosen because they are routinely charted, and no statistical selection was done. One value per hour was used for each of the 43 inputs (eTable 3). Electronic health record data from all sites were converted to hourly patient tables by binning measurements into hours, taking their mean value, and projecting them onto physiologically plausible ranges [23]. Glasgow Coma Scale scores were those charted during routine care. Information on the raters was not available. The same steps were used for all patients. We did not examine whether data quality or missing data differed by age, sex, or race.

*Missing values and scaling.* Missing values were carried forward, then filled with the median across training patients of each patient's median. Inputs were standardized with the training mean and SD, and limited to -5 to 5. Each network used the medians and scaling of its own three training folds at MIMIC-IV. The same steps would apply in use, except on a real-time basis. The index needs no manual input, and no user interface was developed or tested.

*Networks.* Each network had 43 inputs, three hidden layers of 16 units (rectified linear activation), and one output. It was trained on single hours labeled with the patient's outcome, using binary cross-entropy, the Adam optimizer, a learning rate of $1 \times 10^{-5}$, and a batch size of 128, without class weights or label smoothing. Training stopped after 20 epochs without improvement in validation-fold accuracy (limit of 200 epochs, never reached), and the best epoch was kept. Network size, batch size, and learning rate came from Bayesian optimization on MIMIC-IV validation folds. Each of four folds was trained with five seeds (0, 1, 14, 42, and 133), giving 20 networks.

*Index scoring.* Each network's output was standardized with the mean and SD of its outputs over its own validation fold. The index was the mean of the 20 standardized outputs, signed so that higher means higher risk. For development patients, only the five networks that did not train on the patient were used. At each landmark time, inputs came from the window from 6 hours before ICU admission to the landmark.

*Other index versions.* For the version with age fixed, age was set to 65 in every patient. For the version blind to age, sex, and weight, age was set to 65 and weight to 79.7 kg (the development median), and scores with sex set to female and to male were averaged. Networks were not retrained.

## Appendix 3: Comparators and Negative Controls

*Window.* Except for the change in SOFA-2, each comparator used data from 6 hours before ICU admission to the landmark.

*SOFA-2.* Each component was the worst value in the window, and a component with no value scored 0. The respiratory component used $PaO_2/FiO_2$, or $SpO_2/FiO_2$ when no $PaO_2/FiO_2$ was available and $SpO_2$ was below 98% (300 or lower, 1 point; 250 or lower, 2; 200 or lower, 3; and 120 or lower, 4). Scores of 3 and 4 required advanced respiratory support. The renal component was 4 with kidney replacement therapy, or with creatinine above 1.2 mg/dL together with potassium of at least 6.0 mmol/L or with pH of 7.20 or lower and bicarbonate of 12 mmol/L or lower. A vasopressor counted only if it ran in the current and previous hour, at the lower of the two rates. Due to data availability, the rules for urine output, sedation and delirium, extracorporeal membrane oxygenation, and mechanical circulatory support were omitted.

*SOFA-2 plus age.* Logistic regression of the outcome on SOFA-2 and a four-knot restricted cubic spline of age (capped at 90), fitted in MIMIC-IV development patients at each landmark and applied unchanged.

*APACHE II.* Worst value of each acute physiology component (missing scored 0), with arterial pH used for acid-base status, or bicarbonate if there was no pH. Age was capped at 90. Creatinine points were doubled if creatinine met the Kidney Disease: Improving Global Outcomes criterion (a rise of at least 0.3 mg/dL within 48 hours or to at least 1.5 times the lowest value in 7 days) in patients without chronic dialysis. Chronic health points came from diagnosis codes of current and earlier admissions, and the code lists are available with the code.

*Lactate and change in SOFA-2.* Lactate was the highest value in the window. Patients without a value were placed below every measured value, and the analysis was repeated among measured patients only. Change in SOFA-2 was the SOFA-2 of the landmark hour minus that of the first ICU hour, and it is not defined at 1 hour.

*Negative controls.* Forty networks were trained as in Appendix 2 with patient labels replaced by a fair coin flip (20 networks) or shuffled between patients (20 networks).

**Appendix 4: Trial Simulation**

*Enrollment.* Patients with SOFA-2 of at least 6 were eligible. The simulated trial enrolled a fixed percentage of the eligible patients (10% to 90%), taking those with the highest score first. The number enrolled was rounded to whole patients, and ties in score were broken at random.

*Event rates. We* set the control arm event rate $p_0$ as the observed rate of death or hospice discharge in the enrolled patients. The treated arm event rate was $p_1 = 0.80\ p_0$, a 20% relative reduction as in an earlier enrichment study [12], or 0.85 $p_0$ in the sensitivity analysis.

*Patients per arm.* Patients per arm were calculated for a two-sided test of two proportions with α = 0.05 and 80% power ($\beta = 0.20$) [S1]:

$$n = \frac{\left(Z_{1-0.5\alpha}\sqrt{2\overline{p}(1-\overline{p})}+Z_{\beta}\sqrt{p_0(1-p_0)+p_1(1-p_1)}\right)^2}{(p_0-p_1)^2} \text{ with } \overline{p} = \frac{(p_0+p_1)}{2}$$

Enrollment and the event rate were recalculated in each bootstrap resample.

*Early deaths.* Patients who died or went to hospice within 24 hours after the landmark were removed, and enrollment was repeated in the remaining patients.

**Appendix 5: Triage Simulation**

*Protocol.* The simulated hospital followed a model policy (March 26, 2020 version) [8]. Patients were divided into four severity bands by SOFA: <6, 6-8, 9-11, and ≥12. Each patient was assigned points for SOFA band and for comorbitities, and healthier patients (fewer points) were treated first. We used only the SOFA points, so patients were served in order of SOFA-2 bands. Within the last band that had places, patients in the youngest age category went first (younger than 41, 41–60, 61–75, and over 75 years), and then places were given by lottery. Due to data availability, we did not give priority to people whose work is vital to the public health response, as the policy does. The number of places, *K*, was the number of patients with SOFA-2 below 12 (mild shortage), 9 (moderate), or 6 (severe), and was the same for every score.

*Comparator scores.* We used the version of the index that was blinded to age, sex, and weight. For each comparator, three cutoff-points were set so that its four bands held the same shares of MIMIC-IV development patients as the SOFA-2 bands at 24 hours (58.5%, 25.7%, 10.6%, and 5.2%). They were then applied unchanged at every site.

*Sensitivity analyses.* Cutoff-points were set again at 6 and 72 hours, within each site, or for SOFA-2 bands of <9, 9-11, 12-14, and ≥15. The 72-hour analysis included patients still in the ICU, with SOFA-2 from the worst values at 48 to 72 hours.

*Survivors excluded.* Survivors excluded per 1,000 patients were

$$E = \frac{1000}{N} \sum (1 - \pi_i), \text{ with } \pi_i = \min\left(1,\ \max\left(0,\ \frac{K - B_i}{m_i}\right)\right),$$

where the sum is over the patients who survived to discharge (not to hospice), $N$ is the number of patients, $\pi_i$ is the probability that patient *i* receives a place, $B_i$ is the number of patients served before that patient's band and age category, and $m_i$ is the number in the same band and age category. *K* was recalculated in each bootstrap resample, and one lottery was drawn and used for every score.

*Group gaps.* *E* was calculated within each group, at the moderate shortage, and compared between female and male patients and between Black and White patients.

**Appendix 6: Statistics and Stability**

*Intervals.* Outcome rates have Wilson 95% CIs. Other intervals were the 2.5th and 97.5th percentiles of 2,000 bootstrap resamples of patients at each site, using the same resamples for every score. Intervals were not adjusted for multiple comparisons.

*Pooling.* Sites were pooled with random-effects models (restricted maximum likelihood, Wald CIs), on the log scale for ratios. The prediction interval used a t distribution with $k - 2$ degrees of freedom ($k$, number of sites; minimum 1). Estimates from fewer than 100 outcomes were not pooled.

*Calibration.* Logistic models of the outcome on each score were fitted in MIMIC-IV development patients and applied at every site. Calibration was also reported after shifting only the intercept at each site, using 5-fold cross-fitting.

*Stability***.** Among patients still at risk at 25 hours, decisions at 23, 24, and 25 hours were compared between adjacent hours with the same random tie-breaks. The enrollment decision was the top 50% of eligible patients at each hour, and the triage decision was a place at the moderate shortage, with *K* fixed at 24 hours and the bands of Appendix 5. The 3-hour average was the mean of the blind index at the hour and the 2 hours before it.

*Software.* Python 3.11.7, PyTorch 2.9.0, NumPy 1.26.4, pandas 2.2.2, SciPy 1.17.0, scikit-learn 1.8.0, and statsmodels 0.14.5.

**eTable 1:** TRIPOD+AI checklist for the reporting of prediction model studies

| Item | Type | Checklist item | Reported on page |
|---|---|---|---|
| **Title** | | | |
| 1 | D;E | Identify the study as developing or evaluating the performance of a multivariable prediction model, the target population, and the outcome to be predicted | 1 (population; the model and outcome are named in the abstract, page 2) |
| **Abstract** | | | |
| 2 | D;E | See TRIPOD+AI for Abstracts checklist | 2–3 |
| **Introduction** | | | |
| 3a | D;E | Explain the healthcare context (including whether diagnostic or prognostic) and rationale for developing or evaluating the prediction model, including references to existing models | 4–5 |
| 3b | D;E | Describe the target population and the intended purpose of the prediction model in the context of the care pathway, including its intended users (eg, healthcare professionals, patients, public) | 4–6 (users not named) |
| 3c | D;E | Describe any known health inequalities between sociodemographic groups | 13–14 |
| 4 | D;E | Specify the study objectives, including whether the study describes the development or validation of a prediction model (or both) | 5–6 |
| **Methods** | | | |
| 5a | D;E | Describe the sources of data separately for the development and evaluation datasets (eg, randomized trial, cohort, routine care or registry data), the rationale for using these data, and representativeness of the data | 6; representativeness: 14 |
| 5b | D;E | Specify the dates of the collected participant data, including start and end of participant accrual; and, if applicable, end of follow-up | 6 (years) |
| 6a | D;E | Specify key elements of the study setting (eg, primary care, secondary care, general population) including the number and location of centers | 2, 6 |
| 6b | D;E | Describe the eligibility criteria for study participants | 6; Appendix 1; eTable 2 |
| 6c | D;E | Give details of any treatments received, and how they were handled during model development or evaluation, if relevant | 7; Table 1 |
| 7 | D;E | Describe any data pre-processing and quality checking, including whether this was similar across relevant sociodemographic groups | 7; Appendix 2 |
| 8a | D;E | Clearly define the outcome that is being predicted and the time horizon, including how and when assessed, the rationale for choosing this outcome, and whether the method of outcome assessment is consistent across sociodemographic groups | 7; Appendix 1 |
| 8b | D;E | If outcome assessment requires subjective interpretation, describe the qualifications and demographic characteristics of the outcome assessors | Not applicable (outcomes came from records) |
| 8c | D;E | Report any actions to blind assessment of the outcome to be predicted | Not applicable (outcomes came from records) |
| 9a | D | Describe the choice of initial predictors (eg, literature, previous models, all available predictors) and any pre-selection of predictors before model building | 7; Appendix 2 |
| 9b | D;E | Clearly define all predictors, including how and when they were measured (and any actions to blind assessment of predictors for the outcome and other predictors) | 7; Appendix 2; eTable 3 |
| 9c | D;E | If predictor measurement requires subjective interpretation, describe the qualifications and demographic characteristics of the predictor assessors | Appendix 2 |
| 10 | D;E | Explain how the study size was arrived at (separately for development and evaluation), and justify that the study size was sufficient to answer the research question. Include details of any sample size calculation | Appendix 1; eTable 2 |
| 11 | D;E | Describe how missing data were handled. Provide reasons for omitting any data | 7; Appendices 2 and 3 |
| 12a | D | Describe how the data were used (eg, for development and evaluation of model performance) in the analysis, including whether the data were partitioned, considering any sample size requirements | 6–7; Appendix 1 |
| 12b | D | Depending on the type of model, describe how predictors were handled in the analyses (functional form, rescaling, transformation, or any standardization) | 7; Appendix 2 |
| 12c | D | Specify the type of model, rationale, all model building steps, including any hyperparameter tuning, and method for internal validation | 7; Appendix 2 |
| 12d | D;E | Describe if and how any heterogeneity in estimates of model parameter values and model performance was handled and quantified across clusters (eg, hospitals, countries). See TRIPOD-Cluster for additional considerations | 10; Appendix 6 |
| 12e | D;E | Specify all measures and plots used (and their rationale) to evaluate model performance (eg, discrimination, calibration, clinical utility) and, if relevant, to compare multiple models | 8–10; Appendices 4–6 |
| 12f | E | Describe any model updating (eg, recalibration) arising from the model evaluation, either overall or for particular sociodemographic groups or settings | Appendix 6 |
| 12g | E | For model evaluation, describe how the model predictions were calculated (eg, formula, code, object, application programming interface) | 7; Appendix 2; code: 18 |
| 13 | D;E | If class imbalance methods were used, state why and how this was done, and any subsequent methods to recalibrate the model or the model predictions | Not used (Appendix 2) |
| 14 | D;E | Describe any approaches that were used to address model fairness and their rationale | 8–9; Appendices 2 and 5 |
| 15 | D | Specify the output of the prediction model (eg, probabilities, classification). Provide details and rationale for any classification and how the thresholds were identified | 7, 9; Appendices 2 and 5 |
| 16 | D;E | Identify any differences between the development and evaluation data in healthcare setting, eligibility criteria, outcome, and predictors | 6–7; Table 1; eTable 3 |
| 17 | D;E | Name the institutional research board or ethics committee that approved the study and describe the participant informed consent or the ethics committee waiver of informed consent | 6 (exempt) |
| **Open science** | | | |
| 18a | D;E | Give the source of funding and the role of the funders for the present study | 18 |
| 18b | D;E | Declare any conflicts of interest and financial disclosures for all authors | 18 |
| 18c | D;E | Indicate where the study protocol can be accessed or state that a protocol was not prepared | 18 |
| 18d | D;E | Provide registration information for the study, including register name and registration number, or state that the study was not registered | 18 |
| 18e | D;E | Provide details of the availability of the study data | 18 |
| 18f | D;E | Provide details of the availability of the analytical code | 18 |
| **Patient and public involvement** | | | |
| 19 | D;E | Provide details of any patient and public involvement during the design, conduct, reporting, interpretation, or dissemination of the study or state no involvement | 18 |
| **Results** | | | |
| 20a | D;E | Describe the flow of participants through the study, including the number of participants with and without the outcome and, if applicable, a summary of the follow-up time. A diagram may be helpful | 10; eTable 2; Table 1 |
| 20b | D;E | Report the characteristics overall and, where applicable, for each data source or setting, including the key dates, key predictors (including demographics), treatments received, sample size, number of outcome events, follow-up time, and amount of missing data. A table may be helpful. Report any differences across key demographic groups | Table 1; eTable 3 |
| 20c | E | For model evaluation, show a comparison with the development data of the distribution of important predictors (demographics, predictors, and outcome) | Table 1; eTable 3 |
| 21 | D;E | Specify the number of participants and outcome events in each analysis (eg, for model development, hyperparameter tuning, model evaluation) | 10; Table 1; eTable 2 |
| 22 | D | Provide details of the full prediction model (eg, formula, code, object, application programming interface) to allow predictions in new individuals and to enable third party evaluation and implementation, including any restrictions to access or reuse (eg, freely available, proprietary) | Appendix 2; code: 18 |
| 23a | D;E | Report model performance estimates with confidence intervals, including for any key subgroups (eg, sociodemographic). Consider plots to aid presentation | 10–12; Figures 1–4; eTables 4–9; eFigure 1 |
| 23b | D;E | If examined, report results of any heterogeneity in model performance across clusters. See TRIPOD-Cluster for additional details | 11; eTable 6 |
| 24 | E | Report the results from any model updating, including the updated model and subsequent performance | 10–11; eTable 5; eFigure 1 |
| **Discussion** | | | |
| 25 | D;E | Give an overall interpretation of the main results, including issues of fairness in the context of the objectives and previous studies | 12–14 |
| 26 | D;E | Discuss any limitations of the study (such as a non-representative sample, sample size, overfitting, missing data) and their effects on any biases, statistical uncertainty, and generalizability | 14 |
| 27a | D | Describe how poor quality or unavailable input data (eg, predictor values) should be assessed and handled when implementing the prediction model | Appendix 2 (missing inputs) |
| 27b | D | Specify whether users will be required to interact in the handling of the input data or use of the model, and what level of expertise is required of users | Appendix 2 (user expertise not assessed) |
| 27c | D;E | Discuss any next steps for future research, with a specific view to applicability and generalizability of the model | 13–14 |

TRIPOD, Transparent Reporting of a multivariable prediction model for Individual Prognosis Or Diagnosis; AI, artificial intelligence. D, items for developing a prediction model; E, items for evaluating a prediction model; D;E, items for both. Pages are those of the main manuscript. Table 1 is a separate file. Adapted from [S2].

**eTable 2:** Patients remaining after each inclusion and exclusion step

| **Step** | MIMIC-IV | Emory | University of Pittsburgh |
|---|---|---|---|
| **Admissions** | | | |
| Met Sepsis-3 inclusion | 46,382 | 11,808 | 4,790 |
| **Excluded, in order** | | | |
| Not in the ICU from 24 hours before to 48 hours after sepsis onset | 7,000 | 2,290 | 634 |
| Exceeded a rate cutoff | 0 | 5 | 0 |
| Died with no recorded fluids or vasopressors | 1,647 | 260 | 165 |
| Validation admission of a person with a development admission | 3,044 | 0 | 0 |
| Later validation admission of the same person | 575 | 520 | 212 |
| **Cohorts** | | | |
| Development | 23,293 | — | — |
| Validation | 10,823 | 8,733 | 3,779 |
| **Alive in hospital** | | | |
| Development at 1 h | 23,293 | — | — |
| Validation at 1 h | 10,821 | 8,733 | 3,779 |
| Development at 6 h | 23,293 | — | — |
| Validation at 6 h | 10,779 | 8,709 | 3,773 |
| Development at 12 h | 23,293 | — | — |
| Validation at 12 h | 10,661 | 8,670 | 3,761 |
| Development at 24 h | 23,267 | — | — |
| Validation at 24 h | 10,349 | 8,523 | 3,716 |

MIMIC-IV, Medical Information Mart for Intensive Care IV. A dash means that the site had no development patients. Development counts are admissions (20,211 people), and validation counts are one admission per person. Died includes discharge to hospice. The rate cutoffs are in Appendix 1. At University of Pittsburgh, the ICU step was applied last, because ICU times were obtained later.

**eTable 3:** Model Inputs

| Variable Name | Type | Units | Description |
|---|---|---|---|
| Age | Demographics | years | Patient age at ICU admission; constant within a trajectory. |
| Body Weight | Demographics | kg | Charted daily body weight. |
| Sex | Demographics | binary | Administrative sex, encoded 1 = male and 0 = female. |
| Diastolic Blood Pressure | Vital Sign | mmHg | Arterial pressure during diastole, measured by either cuff or line. |
| Heart Rate | Vital Sign | beats/min | Heart rate. |
| Mean Arterial Pressure | Vital Sign | mmHg | Mean arterial pressure, measured by either cuff or line. |
| Peripheral Oxygen Saturation ($SpO_2$) | Vital Sign | % | Peripheral oxygen saturation measured by pulse oximetry. |
| Respiratory Rate | Vital Sign | breaths/min | Spontaneous (unassisted) respiratory rate. |
| Systolic Blood Pressure | Vital Sign | mmHg | Arterial pressure during systole, measured by either cuff or line. |
| Temperature | Vital Sign | °C | Measurement of the body's core temperature. |
| End-Tidal $CO_2$ | Ventilation | mmHg | Partial pressure of $CO_2$ at the end of an exhaled breath. |
| Fraction of Inspired Oxygen ($FiO_2$) | Ventilation | fraction | Fraction of inspired oxygen, 0.21 (room air) to 1.0. |
| Glasgow Coma Scale (Total) | Neurologic Score | points | Glasgow Coma Scale, 3-15; assesses level of consciousness. |
| Alanine Aminotransferase (ALT) | Laboratory Test | U/L | Hepatocellular enzyme indicating liver injury. |
| Albumin | Laboratory Test | g/dL | Serum albumin; marker of hepatic function. |
| Alkaline Phosphatase | Laboratory Test | U/L | Enzyme elevated in cholestasis and bone turnover. |
| Aspartate Aminotransferase (AST) | Laboratory Test | U/L | Enzyme used with ALT to assess hepatocellular injury. |
| Base Excess | Laboratory Test | mmol/L | Excess or deficit of base in the blood. |
| Bicarbonate ($HCO_3^-$) | Laboratory Test | mmol/L | Serum bicarbonate concentration. |
| Bilirubin, Direct | Laboratory Test | mg/dL | Conjugated bilirubin fraction. |
| Bilirubin, Total | Laboratory Test | mg/dL | Total bilirubin, indicating liver function and hemolysis. |
| Blood Urea Nitrogen (BUN) | Laboratory Test | mg/dL | Urea nitrogen level, reflects renal function. |
| Calcium, Ionized | Laboratory Test | mmol/L | Physiologically active free calcium fraction. |
| Calcium, Total | Laboratory Test | mg/dL | Total serum calcium. |
| Chloride | Laboratory Test | mmol/L | Serum chloride concentration. |
| Creatinine | Laboratory Test | mg/dL | Serum creatinine, an indicator of renal function. |
| Fibrinogen | Laboratory Test | mg/dL | Clotting factor consumed in disseminated intravascular coagulation. |
| Glucose | Laboratory Test | mg/dL | Serum glucose concentration. |
| Hematocrit | Laboratory Test | % | Volume fraction of red blood cells in blood. |
| Hemoglobin | Laboratory Test | g/dL | Hemoglobin concentration; marker of oxygen capacity. |
| Lactate | Laboratory Test | mmol/L | Serum lactate, elevated in hypoxia and sepsis. |
| Lactate Dehydrogenase (LDH) | Laboratory Test | U/L | Enzyme indicating tissue damage or hemolysis. |
| Magnesium | Laboratory Test | mg/dL | Serum magnesium concentration. |
| Partial Pressure of Carbon Dioxide ($PaCO_2$) | Laboratory Test | mmHg | Partial pressure of carbon dioxide in arterial blood. |
| Partial Pressure of Oxygen ($PaO_2$) | Laboratory Test | mmHg | Partial pressure of oxygen in arterial blood. |
| Partial Thromboplastin Time (PTT) | Laboratory Test | seconds | Clotting time of the intrinsic coagulation pathway. |
| pH | Laboratory Test | unitless | Arterial blood pH; acidity or alkalinity of the blood. |
| Phosphorus | Laboratory Test | mg/dL | Serum phosphate concentration. |
| Platelet Count | Laboratory Test | $\times 10^3/\mu L$ | Platelet count in the blood. |
| Potassium | Laboratory Test | mmol/L | Serum potassium concentration. |
| Sodium | Laboratory Test | mmol/L | Serum sodium concentration. |
| Troponin | Laboratory Test | ng/mL | Cardiac troponin; marker of myocardial injury. |
| White Blood Cell Count | Laboratory Test | $\times 10^3/\mu L$ | Leukocyte count; marker of infection and inflammation. |

**eTable 4:** Separation of patients by outcome for each score, time, and site

| Score | MIMIC-IV | Emory | University of Pittsburgh | Pooled |
|---|---|---|---|---|
| **Tied-pair AUROC, 1 hour** | | | | |
| Index | 0.711 (0.691 to 0.729) | 0.677 (0.663 to 0.692) | 0.655 (0.631 to 0.679) | 0.682 (0.651 to 0.712) |
| SOFA-2 plus age | 0.579 (0.558 to 0.601) | 0.613 (0.599 to 0.627) | 0.599 (0.574 to 0.624) | 0.598 (0.578 to 0.619) |
| APACHE II | 0.622 (0.601 to 0.642) | 0.581 (0.566 to 0.597) | 0.594 (0.570 to 0.618) | 0.599 (0.574 to 0.623) |
| Lactate | 0.475 (0.455 to 0.496) | 0.537 (0.523 to 0.551) | 0.509 (0.486 to 0.531) | 0.507 (0.472 to 0.543) |
| Range of 40 negative controls | 0.402 to 0.596 | 0.448 to 0.561 | 0.448 to 0.554 | — |
| **Tied-pair AUROC, 6 hours** | | | | |
| Index | 0.763 (0.747 to 0.779) | 0.707 (0.692 to 0.721) | 0.692 (0.670 to 0.714) | 0.721 (0.679 to 0.763) |
| SOFA-2 plus age | 0.591 (0.573 to 0.609) | 0.628 (0.613 to 0.644) | 0.625 (0.603 to 0.646) | 0.614 (0.591 to 0.638) |
| APACHE II | 0.634 (0.617 to 0.652) | 0.579 (0.564 to 0.594) | 0.580 (0.555 to 0.603) | 0.598 (0.562 to 0.634) |
| Lactate | 0.528 (0.509 to 0.548) | 0.552 (0.536 to 0.568) | 0.553 (0.529 to 0.577) | 0.544 (0.528 to 0.560) |
| Change in SOFA-2 | 0.484 (0.468 to 0.502) | 0.533 (0.518 to 0.548) | 0.506 (0.484 to 0.530) | 0.508 (0.479 to 0.537) |
| Range of 40 negative controls | 0.441 to 0.565 | 0.453 to 0.587 | 0.430 to 0.557 | — |
| **Tied-pair AUROC, 12 hours** | | | | |
| Index | 0.788 (0.773 to 0.803) | 0.713 (0.699 to 0.727) | 0.713 (0.690 to 0.734) | 0.738 (0.689 to 0.788) |
| SOFA-2 plus age | 0.596 (0.577 to 0.616) | 0.626 (0.611 to 0.642) | 0.628 (0.607 to 0.651) | 0.617 (0.596 to 0.637) |
| APACHE II | 0.642 (0.623 to 0.660) | 0.585 (0.569 to 0.600) | 0.578 (0.553 to 0.601) | 0.602 (0.562 to 0.642) |
| Lactate | 0.537 (0.516 to 0.557) | 0.549 (0.532 to 0.566) | 0.551 (0.527 to 0.574) | 0.546 (0.535 to 0.557) |
| Change in SOFA-2 | 0.505 (0.485 to 0.525) | 0.548 (0.532 to 0.563) | 0.532 (0.509 to 0.557) | 0.529 (0.503 to 0.554) |
| Range of 40 negative controls | 0.431 to 0.582 | 0.459 to 0.590 | 0.438 to 0.570 | — |
| **Tied-pair AUROC, 24 hours** | | | | |
| Index | 0.789 (0.773 to 0.804) | 0.727 (0.712 to 0.742) | 0.711 (0.688 to 0.732) | 0.742 (0.696 to 0.789) |
| SOFA-2 plus age | 0.600 (0.580 to 0.620) | 0.633 (0.617 to 0.648) | 0.638 (0.615 to 0.660) | 0.624 (0.601 to 0.646) |
| APACHE II | 0.645 (0.626 to 0.664) | 0.590 (0.574 to 0.607) | 0.569 (0.543 to 0.594) | 0.602 (0.558 to 0.646) |
| Lactate | 0.517 (0.496 to 0.539) | 0.552 (0.535 to 0.569) | 0.540 (0.513 to 0.564) | 0.537 (0.516 to 0.558) |
| Change in SOFA-2 | 0.524 (0.503 to 0.546) | 0.566 (0.549 to 0.581) | 0.553 (0.529 to 0.576) | 0.548 (0.524 to 0.573) |
| Range of 40 negative controls | 0.446 to 0.579 | 0.449 to 0.602 | 0.450 to 0.574 | — |
| **AUROC among all patients, 1 hour** | | | | |
| Index | 0.763 (0.749 to 0.776) | 0.711 (0.698 to 0.723) | 0.697 (0.679 to 0.716) | 0.724 (0.685 to 0.763) |
| SOFA-2 | 0.649 (0.634 to 0.665) | 0.611 (0.597 to 0.625) | 0.610 (0.589 to 0.631) | 0.624 (0.598 to 0.649) |

| Score | MIMIC-IV | Emory | University of Pittsburgh | Pooled |
|---|---|---|---|---|
| SOFA-2 plus age | 0.675 (0.660 to 0.691) | 0.656 (0.643 to 0.669) | 0.653 (0.632 to 0.673) | 0.662 (0.648 to 0.676) |
| APACHE II | 0.691 (0.678 to 0.705) | 0.627 (0.614 to 0.640) | 0.638 (0.617 to 0.657) | 0.652 (0.613 to 0.691) |
| **AUROC among all patients, 6 hours** | | | | |
| Index | 0.814 (0.803 to 0.826) | 0.748 (0.736 to 0.761) | 0.741 (0.723 to 0.759) | 0.768 (0.723 to 0.814) |
| SOFA-2 | 0.670 (0.656 to 0.686) | 0.655 (0.641 to 0.669) | 0.661 (0.641 to 0.681) | 0.662 (0.652 to 0.672) |
| SOFA-2 plus age | 0.695 (0.680 to 0.711) | 0.693 (0.680 to 0.706) | 0.695 (0.676 to 0.713) | 0.694 (0.686 to 0.703) |
| APACHE II | 0.705 (0.691 to 0.721) | 0.656 (0.643 to 0.669) | 0.667 (0.647 to 0.687) | 0.676 (0.647 to 0.706) |
| **AUROC among all patients, 12 hours** | | | | |
| Index | 0.833 (0.821 to 0.844) | 0.756 (0.744 to 0.768) | 0.762 (0.745 to 0.780) | 0.784 (0.736 to 0.833) |
| SOFA-2 | 0.678 (0.661 to 0.695) | 0.669 (0.655 to 0.683) | 0.679 (0.658 to 0.699) | 0.674 (0.665 to 0.683) |
| SOFA-2 plus age | 0.703 (0.687 to 0.718) | 0.704 (0.691 to 0.717) | 0.711 (0.691 to 0.730) | 0.705 (0.696 to 0.714) |
| APACHE II | 0.714 (0.699 to 0.728) | 0.674 (0.661 to 0.687) | 0.680 (0.660 to 0.700) | 0.689 (0.665 to 0.714) |
| **AUROC among all patients, 24 hours** | | | | |
| Index | 0.823 (0.810 to 0.836) | 0.770 (0.757 to 0.782) | 0.764 (0.746 to 0.781) | 0.786 (0.748 to 0.823) |
| SOFA-2 | 0.666 (0.648 to 0.684) | 0.677 (0.664 to 0.692) | 0.687 (0.668 to 0.707) | 0.676 (0.667 to 0.686) |
| SOFA-2 plus age | 0.692 (0.675 to 0.709) | 0.713 (0.700 to 0.726) | 0.721 (0.703 to 0.740) | 0.709 (0.693 to 0.725) |
| APACHE II | 0.704 (0.690 to 0.721) | 0.688 (0.675 to 0.701) | 0.682 (0.661 to 0.702) | 0.692 (0.679 to 0.705) |

AUROC, area under the receiver operating characteristic curve; SOFA-2, Sequential Organ Failure Assessment 2; APACHE II, Acute Physiology and Chronic Health Evaluation II; CI, confidence interval. Values are estimates (95% CI). Within pairs with the same SOFA-2, SOFA-2 itself has an AUROC of 0.5. Change in SOFA-2 is not defined at 1 hour. Negative controls are networks trained on coin-flip or shuffled labels, and their range was not pooled.

**eTable 5:** Calibration at 24 hours

| Score | MIMIC-IV | Emory | University of Pittsburgh |
|---|---|---|---|
| **Calibration in the large, before site update** | | | |
| Index | −0.15 | 0.50 | 0.66 |
| SOFA-2 | −0.23 | 0.47 | 0.43 |
| SOFA-2 plus age | −0.27 | 0.52 | 0.52 |
| APACHE II | −0.21 | 0.41 | 0.39 |
| **Calibration in the large, after site update** | | | |
| Index | 0.00 | 0.00 | 0.00 |
| SOFA-2 | 0.00 | 0.00 | 0.00 |
| SOFA-2 plus age | 0.00 | 0.00 | 0.00 |
| APACHE II | 0.00 | 0.00 | 0.00 |
| **Calibration slope, before site update** | | | |
| Index | 1.12 | 0.92 | 0.82 |
| SOFA-2 | 0.99 | 0.89 | 0.90 |
| SOFA-2 plus age | 0.99 | 0.94 | 0.98 |
| APACHE II | 1.04 | 0.90 | 0.86 |
| **Calibration slope, after site update** | | | |
| Index | 1.11 | 0.92 | 0.82 |
| SOFA-2 | 0.99 | 0.89 | 0.89 |
| SOFA-2 plus age | 0.98 | 0.94 | 0.98 |
| APACHE II | 1.04 | 0.90 | 0.85 |
| **Integrated calibration index, before site update** | | | |
| Index | 0.013 | 0.061 | 0.080 |
| SOFA-2 | 0.024 | 0.067 | 0.065 |
| SOFA-2 plus age | 0.027 | 0.071 | 0.075 |
| APACHE II | 0.022 | 0.058 | 0.060 |
| **Integrated calibration index, after site update** | | | |
| Index | 0.009 | 0.012 | 0.025 |
| SOFA-2 | 0.010 | 0.012 | 0.009 |
| SOFA-2 plus age | 0.011 | 0.007 | 0.008 |
| APACHE II | 0.008 | 0.009 | 0.017 |

MIMIC-IV, Medical Information Mart for Intensive Care IV; SOFA-2, Sequential Organ Failure Assessment 2; APACHE II, Acute Physiology and Chronic Health Evaluation II; The link from score to risk was fitted in MIMIC-IV development patients. The site update shifts the intercept only (5-fold cross-fitted). Perfect values are 0 for calibration in the large, 1 for the slope, and 0 for the integrated calibration index. Positive calibration in the large means that predicted risk was too low.

**eTable 6:** New hospital pooled differences and prediction intervals

| Comparison | Pooled (95% CI) | 95% prediction interval | $I^2$, % |
|---|---|---|---|
| **Tied-pair AUROC, index minus APACHE II** | | | |
| 1 hour | 0.084 (0.065 to 0.104) | −0.123 to 0.291 | 56 |
| 6 hours | 0.125 (0.113 to 0.138) | 0.045 to 0.206 | 0 |
| 12 hours | 0.136 (0.124 to 0.149) | 0.054 to 0.219 | 0 |
| 24 hours | 0.140 (0.127 to 0.153) | 0.054 to 0.225 | 0 |
| 24 hours, age fixed at 65 | 0.115 (0.098 to 0.132) | −0.047 to 0.278 | 32 |
| 24 hours, blind to age, sex, and weight | 0.107 (0.093 to 0.121) | 0.016 to 0.198 | 0 |
| 24 hours, female patients | 0.140 (0.120 to 0.160) | 0.010 to 0.270 | 0 |
| 24 hours, male patients | 0.139 (0.122 to 0.157) | 0.025 to 0.254 | 0 |
| 24 hours, Black patients | 0.163 (0.135 to 0.190) | −0.016 to 0.341 | 0 |
| 24 hours, White patients | 0.129 (0.105 to 0.154) | −0.122 to 0.380 | 50 |
| **Tied-pair AUROC, index minus SOFA-2 plus age** | | | |
| 1 hour | 0.084 (0.038 to 0.131) | −0.504 to 0.673 | 93 |
| 6 hours | 0.106 (0.041 to 0.171) | −0.724 to 0.936 | 97 |
| 12 hours | 0.121 (0.052 to 0.191) | −0.773 to 1.016 | 97 |
| 24 hours | 0.119 (0.049 to 0.188) | −0.768 to 1.005 | 97 |
| 24 hours, age fixed at 65 | 0.093 (0.010 to 0.176) | −0.970 to 1.155 | 96 |
| 24 hours, blind to age, sex, and weight | 0.085 (0.008 to 0.163) | −0.903 to 1.074 | 96 |
| 24 hours, female patients | 0.124 (0.065 to 0.182) | −0.603 to 0.850 | 88 |
| 24 hours, male patients | 0.114 (0.034 to 0.194) | −0.904 to 1.131 | 96 |
| 24 hours, Black patients | 0.104 (0.080 to 0.128) | −0.052 to 0.260 | 34 |
| 24 hours, White patients | 0.113 (0.039 to 0.187) | −0.833 to 1.059 | 95 |
| **Tied-pair AUROC, index minus Lactate** | | | |
| 1 hour | 0.174 (0.114 to 0.234) | −0.586 to 0.934 | 95 |
| 6 hours | 0.177 (0.119 to 0.234) | −0.553 to 0.906 | 95 |
| 12 hours | 0.193 (0.135 to 0.251) | −0.545 to 0.930 | 95 |
| 24 hours | 0.206 (0.141 to 0.270) | −0.616 to 1.027 | 95 |
| 24 hours, age fixed at 65 | 0.180 (0.102 to 0.258) | −0.821 to 1.181 | 97 |
| 24 hours, blind to age, sex, and weight | 0.172 (0.099 to 0.246) | −0.761 to 1.106 | 96 |
| 24 hours, female patients | 0.199 (0.153 to 0.245) | −0.348 to 0.746 | 77 |
| 24 hours, male patients | 0.207 (0.131 to 0.284) | −0.761 to 1.176 | 94 |
| 24 hours, Black patients | 0.182 (0.156 to 0.208) | 0.012 to 0.352 | 0 |
| 24 hours, White patients | 0.208 (0.128 to 0.289) | −0.813 to 1.229 | 95 |
| **Patients per arm, index divided by SOFA-2 alone, 50% enrolled** | | | |
| 1 hour | 0.70 (0.65 to 0.76) | 0.42 to 1.19 | 0 |
| 6 hours | 0.73 (0.67 to 0.79) | 0.30 to 1.80 | 61 |
| 12 hours | 0.73 (0.70 to 0.77) | 0.50 to 1.06 | 3 |
| 24 hours | 0.73 (0.68 to 0.79) | 0.29 to 1.86 | 72 |

| Comparison | Pooled (95% CI) | 95% prediction interval | $I^2$, % |
|---|---|---|---|
| 24 hours, death alone | 0.83 (0.71 to 0.96) | 0.13 to 5.36 | 93 |
| 24 hours, 15% relative reduction | 0.73 (0.67 to 0.79) | 0.29 to 1.82 | 71 |
| 24 hours, without early deaths | 0.73 (0.67 to 0.79) | 0.27 to 1.96 | 73 |
| **Survivors excluded per 1,000, index minus SOFA-2** | | | |
| Mild shortage | −8.2 (−11.7 to −4.6) | −48.9 to 32.5 | 75 |
| Moderate shortage | −15.5 (−17.9 to −13.1) | −31.1 to 0.0 | 0 |
| Severe shortage | −17.6 (−20.3 to −14.9) | −35.1 to −0.1 | 0 |
| **Survivors excluded per 1,000, index minus SOFA-2 plus age** | | | |
| Mild shortage | −7.0 (−8.8 to −5.2) | −20.5 to 6.5 | 0 |
| Moderate shortage | −11.4 (−15.2 to −7.6) | −49.6 to 26.8 | 48 |
| Severe shortage | −11.8 (−17.9 to −5.6) | −84.1 to 60.6 | 80 |
| **Survivors excluded per 1,000, index minus APACHE II** | | | |
| Mild shortage | −11.6 (−15.0 to −8.2) | −48.8 to 25.5 | 66 |
| Moderate shortage | −17.4 (−20.2 to −14.6) | −38.1 to 3.3 | 12 |
| Severe shortage | −14.6 (−17.2 to −11.9) | −31.8 to 2.7 | 0 |

AUROC, area under the receiver operating characteristic curve; SOFA-2, Sequential Organ Failure Assessment 2; APACHE II, Acute Physiology and Chronic Health Evaluation II; CI, confidence interval. Pooled over three sites with random-effects models. The prediction interval is the range expected at a new hospital and uses a t distribution with one degree of freedom, because there are three sites. The survivors excluded use the index blind to age, sex, and weight, at 24 hours. Mortality at 28 and 90 days was available only at MIMIC-IV, so it was not pooled.

**eTable 7:** Patients enrolled by each method (Trial Simulation)

| Characteristic | SOFA-2 alone | | | Index | | | APACHE II | | |
|---|---|---|---|---|---|---|---|---|---|
| | Enrolled | Not enrolled | SMD | Enrolled | Not enrolled | SMD | Enrolled | Not enrolled | SMD |
| **MIMIC-IV (3,608 eligible)** | | | | | | | | | |
| Age (y), median | 65 | 68 | −0.14 | 70 | 64 | 0.38 | 70 | 64 | 0.33 |
| Female, % | 35.6 | 38.0 | −0.05 | 41.1 | 32.5 | 0.18 | 40.0 | 33.5 | 0.13 |
| Black, % | 8.2 | 8.8 | −0.02 | 9.5 | 7.4 | 0.08 | 9.3 | 7.7 | 0.06 |
| White, % | 63.7 | 69.0 | −0.11 | 63.3 | 69.4 | −0.13 | 64.6 | 68.1 | −0.08 |
| SOFA-2, median | 9 | 6 | 2.05 | 8 | 7 | 0.60 | 8 | 7 | 0.79 |
| Vasopressors, % | 67.4 | 53.5 | 0.29 | 56.8 | 64.0 | −0.15 | 59.9 | 60.9 | −0.02 |
| Invasive mechanical ventilation, % | 75.7 | 58.0 | 0.38 | 58.3 | 75.4 | −0.37 | 67.0 | 66.8 | 0.00 |
| Kidney replacement therapy, % | 11.4 | 4.0 | 0.28 | 11.0 | 4.4 | 0.25 | 11.6 | 3.8 | 0.30 |
| Enrolled by the index and by SOFA-2 alone, % | | | | 60.1 | | | | | |
| **Emory (3,918 eligible)** | | | | | | | | | |
| Age (y), median | 64 | 64 | −0.02 | 67 | 60 | 0.53 | 66 | 61 | 0.32 |
| Female, % | 40.4 | 45.9 | −0.11 | 43.6 | 42.7 | 0.02 | 42.8 | 43.5 | −0.01 |
| Black, % | 44.4 | 46.2 | −0.04 | 43.8 | 46.8 | −0.06 | 46.4 | 44.3 | 0.04 |
| White, % | 44.2 | 44.5 | −0.01 | 44.9 | 43.8 | 0.02 | 42.8 | 45.8 | −0.06 |
| SOFA-2, median | 10 | 7 | 2.33 | 9 | 7 | 0.65 | 10 | 7 | 0.95 |
| Vasopressors, % | 77.5 | 39.3 | 0.84 | 65.5 | 51.3 | 0.29 | 67.6 | 49.2 | 0.38 |
| Invasive mechanical ventilation, % | 69.0 | 39.9 | 0.61 | 51.7 | 57.2 | −0.11 | 67.7 | 41.1 | 0.55 |
| Kidney replacement therapy, % | 12.8 | 8.1 | 0.15 | 10.0 | 10.9 | −0.03 | 10.5 | 10.4 | 0.00 |
| Enrolled by the index and by SOFA-2 alone, % | | | | 62.3 | | | | | |

| Characteristic | SOFA-2 alone | | | Index | | | APACHE II | | |
|---|---|---|---|---|---|---|---|---|---|
| | Enrolled | Not enrolled | SMD | Enrolled | Not enrolled | SMD | Enrolled | Not enrolled | SMD |
| **University of Pittsburgh (2,108 eligible)** | | | | | | | | | |
| Age (y), median | 60 | 62 | −0.08 | 64 | 58 | 0.51 | 63 | 59 | 0.26 |
| Female, % | 40.4 | 40.7 | −0.01 | 43.4 | 37.8 | 0.11 | 41.2 | 39.9 | 0.03 |
| Black, % | 13.3 | 13.0 | 0.01 | 13.8 | 12.5 | 0.04 | 14.1 | 12.1 | 0.06 |
| White, % | 72.7 | 74.1 | −0.03 | 74.3 | 72.5 | 0.04 | 72.9 | 73.9 | −0.02 |
| SOFA-2, median | 11 | 7 | 2.32 | 10 | 7 | 0.83 | 10 | 7 | 1.03 |
| Vasopressors, % | 74.0 | 36.1 | 0.82 | 68.6 | 41.6 | 0.56 | 69.5 | 40.6 | 0.61 |
| Invasive mechanical ventilation, % | 86.1 | 68.6 | 0.43 | 73.9 | 80.8 | −0.17 | 88.4 | 66.3 | 0.55 |
| Kidney replacement therapy, % | 14.2 | 5.6 | 0.29 | 12.0 | 7.8 | 0.14 | 13.8 | 6.1 | 0.26 |
| Enrolled by the index and by SOFA-2 alone, % | | | | 65.3 | | | | | |

SMD, standardized mean difference; SOFA-2, Sequential Organ Failure Assessment 2; APACHE II, Acute Physiology and Chronic Health Evaluation II. Eligible patients had SOFA-2 of at least 6 at 24 hours, and each method enrolled 50% of them. Organ support is any use from 6 hours before ICU admission to 24 hours. The last row of each site is the share of enrolled patients whom both the index and SOFA-2 alone enroll.

**eTable 8:** Survivors excluded per patient group (Triage Simulation)

| Protocol | MIMIC-IV | Emory | University of Pittsburgh | Pooled |
|---|---|---|---|---|
| **All patients** | | | | |
| SOFA-2 | 78.8 (73.9 to 84.0) | 119.6 (112.9 to 126.0) | 152.3 (141.0 to 163.3) | 116.7 (75.1 to 158.3) |
| Index, age tie-breaker | 63.3 (58.1 to 68.7) | 103.5 (96.9 to 110.3) | 137.9 (127.3 to 148.8) | 101.4 (59.2 to 143.6) |
| Index, lottery only | 63.8 (58.2 to 69.4) | 104.7 (97.5 to 112.0) | 142.5 (131.1 to 154.2) | 103.4 (58.9 to 147.9) |
| **Female** | | | | |
| SOFA-2 | 61.1 (54.1 to 68.5) | 104.8 (95.5 to 114.1) | 140.2 (124.0 to 156.5) | 101.5 (56.8 to 146.2) |
| Index, age tie-breaker | 62.5 (54.5 to 70.7) | 104.0 (93.6 to 114.4) | 145.5 (128.9 to 161.7) | 103.5 (56.7 to 150.4) |
| Index, lottery only | 62.0 (54.3 to 69.8) | 104.3 (93.8 to 115.0) | 147.9 (131.5 to 164.6) | 104.3 (55.8 to 152.7) |
| **Male** | | | | |
| SOFA-2 | 91.5 (84.5 to 98.5) | 132.3 (122.5 to 141.7) | 161.5 (145.7 to 177.3) | 128.0 (88.3 to 167.7) |
| Index, age tie-breaker | 63.9 (57.7 to 70.5) | 103.1 (94.9 to 111.9) | 132.1 (117.6 to 146.3) | 99.3 (60.7 to 138.0) |
| Index, lottery only | 65.0 (58.4 to 72.4) | 105.0 (96.1 to 114.2) | 138.5 (123.1 to 154.2) | 102.4 (60.9 to 143.9) |
| **Black** | | | | |
| SOFA-2 | 83.7 (66.4 to 102.9) | 125.6 (115.0 to 135.7) | 163.8 (130.9 to 197.9) | 122.7 (78.9 to 166.5) |
| Index, age tie-breaker | 78.3 (61.5 to 97.1) | 100.4 (90.8 to 110.1) | 141.4 (112.1 to 172.5) | 104.7 (70.8 to 138.5) |
| Index, lottery only | 82.4 (65.2 to 102.2) | 103.5 (93.1 to 114.3) | 163.5 (131.5 to 198.7) | 114.5 (69.1 to 159.8) |
| **White** | | | | |
| SOFA-2 | 74.8 (69.1 to 81.0) | 110.5 (101.6 to 120.4) | 150.1 (137.2 to 163.3) | 111.5 (68.9 to 154.1) |
| Index, age tie-breaker | 61.8 (55.3 to 68.3) | 105.5 (95.6 to 115.2) | 138.9 (125.6 to 152.4) | 101.7 (58.0 to 145.5) |
| Index, lottery only | 61.2 (55.1 to 67.4) | 104.7 (95.3 to 114.3) | 141.5 (128.5 to 154.5) | 102.2 (56.7 to 147.6) |
| **Age 18–40 y** | | | | |
| SOFA-2 | 71.3 (54.0 to 89.9) | 119.4 (99.3 to 139.3) | 213.9 (178.8 to 248.6) | — |
| Index, age tie-breaker | 83.3 (64.6 to 102.5) | 130.3 (110.2 to 151.5) | 152.0 (124.0 to 185.4) | — |
| Index, lottery only | 90.6 (71.0 to 110.3) | 138.0 (117.6 to 160.5) | 193.0 (161.4 to 227.9) | — |
| **Age 41–60 y** | | | | |
| SOFA-2 | 98.0 (87.1 to 109.8) | 145.4 (131.7 to 158.8) | 186.3 (163.2 to 209.7) | 142.4 (92.7 to 192.2) |
| Index, age tie-breaker | 64.3 (55.4 to 73.2) | 109.2 (96.8 to 121.5) | 124.2 (104.8 to 144.1) | 98.5 (63.0 to 134.0) |
| Index, lottery only | 71.6 (61.7 to 81.9) | 117.1 (104.3 to 129.9) | 163.6 (141.4 to 185.4) | 116.7 (64.9 to 168.4) |

| Protocol | MIMIC-IV | Emory | University of Pittsburgh | Pooled |
|---|---|---|---|---|
| **Age 61–75 y** | | | | |
| SOFA-2 | 82.3 (74.1 to 91.4) | 116.4 (105.3 to 127.9) | 131.6 (114.4 to 149.7) | 109.4 (80.8 to 138.0) |
| Index, age tie-breaker | 50.4 (43.4 to 57.1) | 89.6 (79.8 to 99.6) | 89.3 (75.1 to 118.1) | 75.5 (49.1 to 101.9) |
| Index, lottery only | 57.7 (49.3 to 65.5) | 97.0 (86.4 to 108.0) | 126.3 (107.9 to 143.3) | 93.0 (54.2 to 131.9) |
| **Age over 75 y** | | | | |
| SOFA-2 | 59.2 (50.9 to 67.6) | 87.8 (75.5 to 100.5) | 87.7 (67.2 to 109.7) | 77.0 (57.1 to 96.8) |
| Index, age tie-breaker | 72.8 (54.5 to 91.4) | 105.0 (76.4 to 135.0) | 255.4 (203.0 to 282.3) | 143.4 (33.8 to 252.9) |
| Index, lottery only | 56.8 (48.2 to 65.7) | 81.3 (68.1 to 95.1) | 99.7 (76.8 to 123.6) | 77.5 (53.4 to 101.6) |

SOFA-2, Sequential Organ Failure Assessment 2; CI, confidence interval. Values are survivors excluded per 1,000 patients in each group (95% CI), at the moderate shortage and 24 hours. The index is the version blind to age, sex, and weight. Lottery only means that age did not break ties. A dash means that the estimate was not pooled, because MIMIC-IV and University of Pittsburgh had fewer than 100 outcomes in this age group. At Emory, ethnicity was not extracted, so its Black and White groups may include Hispanic patients.

**eTable 9:** Changes in decisions between adjacent hours

| Score | 23 to 24 hours, % changed (95% CI) | 24 to 25 hours, % changed (95% CI) |
|---|---|---|
| **Trial enrollment** | | |
| SOFA-2 alone | 0.5 (0.2 to 0.8) | 0.5 (0.1 to 0.8) |
| Index | 2.7 (2.1 to 3.3) | 2.8 (2.1 to 3.5) |
| Index blind to age, sex, and weight | 2.9 (2.4 to 3.3) | 3.4 (2.6 to 4.2) |
| Blind index, 3-hour average | 1.6 (1.2 to 2.1) | 1.5 (1.1 to 1.9) |
| **Triage, moderate shortage** | | |
| SOFA-2 | 0.6 (0.5 to 0.8) | 0.6 (0.3 to 0.9) |
| Index blind to age, sex, and weight | 3.8 (2.3 to 5.2) | 3.9 (2.3 to 5.4) |
| Blind index, 3-hour average | 1.9 (1.3 to 2.5) | 2.0 (1.1 to 3.0) |

SOFA-2, Sequential Organ Failure Assessment 2; CI, confidence interval. Values are pooled over the three sites. Enrollment is the top 50% of eligible patients at each hour. Triage is a place at the moderate shortage, with the index blind to age, sex, and weight. The 3-hour average is the mean of the blind index at the hour and the 2 hours before it.

**eTable 10:** Death or hospice discharge by lactate band

| SOFA-2 | Lactate ≤2.0 mmol/L | Lactate >2.0 to 4.0 mmol/L | Lactate >4.0 mmol/L | Not measured |
|---|---|---|---|---|
| 0–1 | 435<br>10.8% (8.2 to 14.1) | 180<br>13.3% (9.1 to 19.1) | 54<br>14.8% (7.7 to 26.6) | 800<br>9.6% (7.8 to 11.9) |
| 2 | 865<br>8.2% (6.6 to 10.2) | 392<br>10.7% (8.0 to 14.2) | 100<br>15.0% (9.3 to 23.3) | 1,111<br>9.0% (7.5 to 10.8) |
| 3 | 1,097<br>8.6% (7.1 to 10.4) | 568<br>10.0% (7.8 to 12.8) | 163<br>15.3% (10.6 to 21.7) | 1,136<br>10.0% (8.4 to 11.9) |
| 4 | 1,188<br>9.3% (7.8 to 11.1) | 723<br>7.9% (6.1 to 10.1) | 212<br>16.5% (12.1 to 22.1) | 1,024<br>9.9% (8.2 to 11.8) |
| 5 | 1,131<br>13.6% (11.7 to 15.7) | 755<br>13.5% (11.3 to 16.1) | 273<br>21.6% (17.1 to 26.9) | 747<br>14.2% (11.9 to 16.9) |
| 6 | 927<br>15.3% (13.1 to 17.8) | 719<br>12.2% (10.0 to 14.8) | 295<br>22.4% (18.0 to 27.5) | 469<br>18.1% (14.9 to 21.9) |
| 7 | 682<br>17.9% (15.2 to 20.9) | 585<br>19.7% (16.6 to 23.1) | 267<br>32.2% (26.9 to 38.0) | 322<br>18.0% (14.2 to 22.6) |
| 8 | 475<br>22.3% (18.8 to 26.3) | 463<br>22.2% (18.7 to 26.3) | 265<br>31.3% (26.0 to 37.1) | 200<br>24.5% (19.1 to 30.9) |
| 9 | 350<br>27.7% (23.3 to 32.6) | 352<br>27.0% (22.6 to 31.9) | 298<br>31.2% (26.2 to 36.7) | 127<br>29.9% (22.6 to 38.4) |
| 10 | 239<br>28.0% (22.7 to 34.0) | 248<br>33.9% (28.3 to 40.0) | 239<br>33.5% (27.8 to 39.7) | 75<br>40.0% (29.7 to 51.3) |
| 11 | 165<br>24.2% (18.3 to 31.3) | 177<br>36.7% (30.0 to 44.0) | 220<br>41.4% (35.1 to 48.0) | 66<br>51.5% (39.7 to 63.2) |
| 12 | 91<br>40.7% (31.1 to 50.9) | 125<br>46.4% (37.9 to 55.1) | 170<br>47.1% (39.7 to 54.5) | 32<br>43.8% (28.2 to 60.7) |
| 13 | 45<br>35.6% (23.2 to 50.2) | 80<br>43.8% (33.4 to 54.7) | 164<br>53.0% (45.4 to 60.5) | 16<br>not shown |
| 14 | 29<br>44.8% (28.4 to 62.5) | 59<br>57.6% (44.9 to 69.4) | 129<br>50.4% (41.9 to 58.9) | 15<br>not shown |
| ≥15 | 44<br>47.7% (33.8 to 62.1) | 57<br>45.6% (33.4 to 58.4) | 323<br>69.3% (64.1 to 74.1) | 30<br>76.7% (59.1 to 88.2) |

SOFA-2, Sequential Organ Failure Assessment 2; CI, confidence interval. The three sites are combined, at 24 hours, by highest lactate within each SOFA-2 value. Each cell gives the number of patients, then the percentage with death or hospice discharge (95% CI). Rates are not shown for fewer than 20 patients. Patients with lactate above 4.0 mmol/L had clearly higher risk than those with 2.0 mmol/L or lower at 7 of 15 SOFA-2 values, where 95% CIs did not overlap.

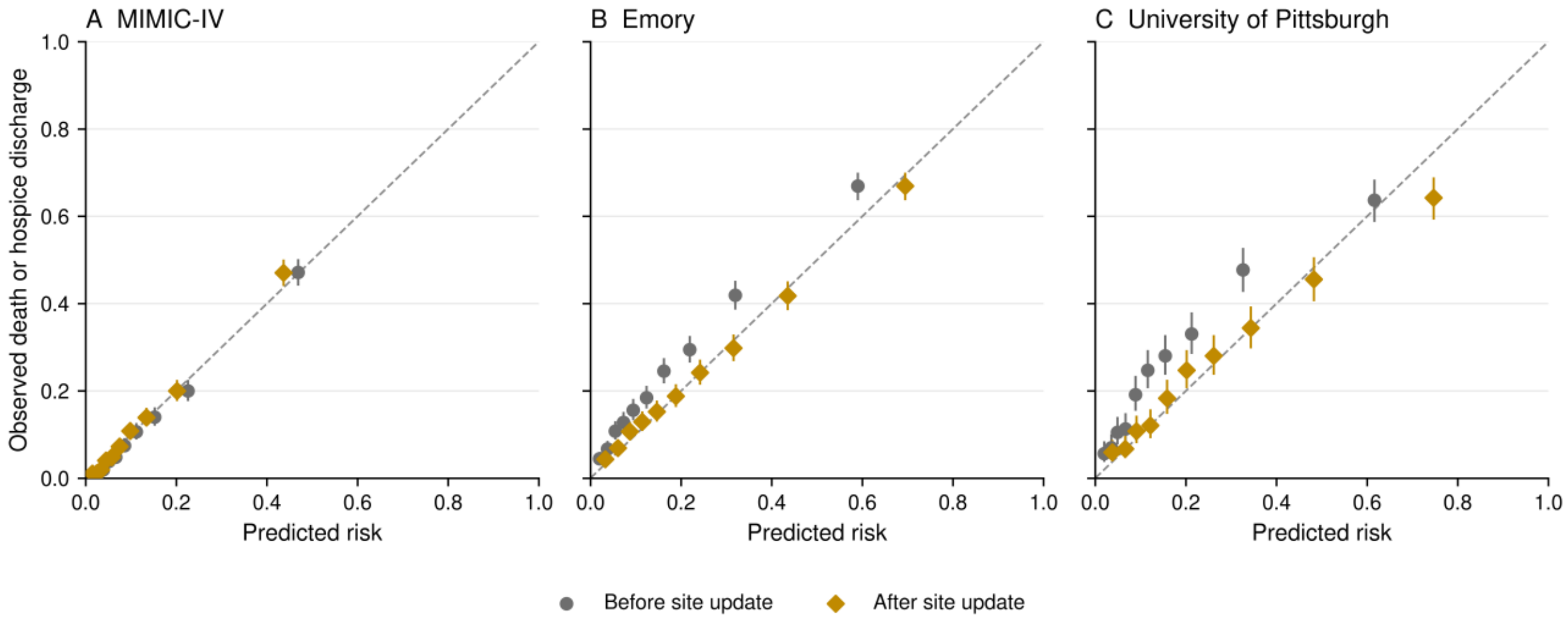

**eFigure 1: Calibration of the index at 24 hours.** Points are at tenths of predicted risk, with the observed proportion with death or hospice discharge and Wilson 95% intervals. The dashed line marks perfect calibration. The link from the index to risk was fitted in MIMIC-IV development patients, before and after updating its intercept at each site. (A) MIMIC-IV, (B) Emory, (C) University of Pittsburgh.

## Supplement References

[1] Fleiss J. L., Tytun A., Ury H. K. A simple approximation for calculating sample sizes for comparing independent proportions. *Biometrics* **36**, 343-346 (1980).

[2] Collins G. S. *et al.* TRIPOD+AI statement: Updated guidance for reporting clinical prediction models that use regression or machine learning methods. *BMJ* **385**, e078378 (2024).